\documentclass[a4paper,12pt,twoside]{article}
\usepackage[utf8]{inputenc}
\usepackage{caption}
\usepackage{subcaption}
\usepackage{todonotes}
\usepackage{tablefootnote}
\usepackage{enumerate}
\usepackage{booktabs, multirow}
\usepackage{amsmath}
\usepackage[english]{babel}
\usepackage{csquotes}
\allowdisplaybreaks[4]

\newcommand{\cS}{ {\mathcal S} }
\usepackage{tikz}
\usetikzlibrary{fit,positioning}
\usetikzlibrary{shapes.geometric, arrows}
\tikzstyle{identity} = [rectangle, rounded corners, minimum width=3cm, minimum height=1cm,text centered, draw=black, fill=red!20]
\tikzstyle{equation} = [rectangle, minimum width=3cm, minimum height=1cm, text centered, draw=black, fill=orange!05]

\tikzstyle{operator} = [circle, text centered, draw=black,inner sep=0pt, text width=8mm, fill=orange!05]

\tikzstyle{leaf} = [circle, text centered, draw=black, inner sep=0pt, text width=8mm, fill=green!05]

\tikzset{
block/.style={
  draw, 
  rectangle, 
  minimum height=1.5cm, 
  minimum width=3cm, align=center
  }, 
line/.style={->,>=latex'}
}

\tikzstyle{arrow} = [thick,->,>=stealth]
\tikzstyle{line} = [thick,>=stealth]
\def\Li{\text{Li}}

\makeatletter
\newcommand\nocaption{%
    \renewcommand\p@subfigure{}
    \renewcommand\thesubfigure{\thefigure\alph{subfigure}}
}
\makeatother
\usepackage{amssymb}
\usepackage{float}
\usepackage{slashed}
\usepackage{mathtools}
\usepackage{physics}
\usepackage{color}
\usepackage[left=2.2cm, right=2.2cm, top=2.2cm,bottom=2.5cm]{geometry}
\usepackage[pdfencoding=auto, psdextra]{hyperref}
\usepackage{graphicx}
\usepackage[backend=bibtex8, sorting=none]{biblatex} 
\definecolor{darkyellow}{rgb}{0.5, 0.5, 0.0}
\definecolor{darkpurple}{rgb}{0.5, 0.2, 0.8}
\definecolor{darkblue}{rgb}{0.0, 0.0, 0.8}
\definecolor{darkgreen}{rgb}{0.0, 0.4, 0.0}
\definecolor{darkred}{rgb}{0.5, 0.0, 0.0}
\usepackage{hyperref}
\hypersetup{
    linktocpage,
     colorlinks,
     citecolor=darkgreen,
     linkcolor= darkgreen,
     urlcolor=darkgreen
}

\title{\textbf{Simplifying Polylogarithms with \\Machine Learning}}

\date{}

\author{Aur\'{e}lien Dersy, Matthew D. Schwartz and Xiaoyuan Zhang\\
{\small 
\href{mailto:adersy@g.harvard.edu}{adersy@g.harvard.edu}, ~
\href{mailto:schwartz@g.harvard.edu}{schwartz@g.harvard.edu}, ~
\href{mailto:xiaoyuanzhang@g.harvard.edu}{xiaoyuanzhang@g.harvard.edu} }\\
\small{Department of Physics, Harvard University, Cambridge MA 02138}\\
\small{The NSF AI Institute for Artificial Intelligence and Fundamental Interactions}
 }
\begin{document}

\maketitle
\begin{abstract}
   Polylogrithmic functions, such as the logarithm or dilogarithm, satisfy a number of algebraic identities. For the logarithm, all the identities follow from the product rule. For the dilogarithm and higher-weight classical polylogarithms, the identities can involve five functions or more. In many calculations relevant to particle physics, complicated combinations of polylogarithms often arise from Feynman integrals. Although the initial expressions resulting from the integration usually simplify, it is often difficult to know which identities to apply and in what order. To address this bottleneck, we explore to what extent machine learning methods can help. We consider both a reinforcement learning approach, where the identities are analogous to moves in a game, and a transformer network approach, where the problem is viewed analogously to a language-translation task. While both methods are effective, the transformer network appears more powerful and holds promise for practical use in symbolic manipulation tasks in mathematical physics. 
\end{abstract}

\section{Introduction}
Polylogrithmic functions, such as $\ln (x)$ and $\Li_2(x)$ naturally appear in many calculations relevant to physical systems. For example, in high energy physics, Feynman integrals which describe the scattering of elementary particles,  polylogarithms are ubiquitous. 
One way to understand the connection between polylogarithms and Feynman integrals is through the method of differential equations \cite{KOTIKOV1991158,Gehrmann:2000zt}, which recasts a Feynman integral in terms of an auxiliary ordinary differential equation which can be integrated multiple times. These nested integrals naturally produce a class of functions which includes the classical polylogarithms.  In recent years, progress has been made in understanding the broader class of functions which can appear in Feynman integrals, such as multiple polylogarithms \cite{Goncharov:1998kja,Borwein:1999js,goncharov1mpl} or elliptic polylogarithms \cite{levin1994elliptic,multipleLevin,multiplebrown}, but the complete characterization of the possible functional forms of Feynman integrals remains an unsolved problem. Nevertheless, even when an integral can be evaluated and expressed only in classical polylogarithms, $\Li_n(x)$, simplifying the initial expression can be challenging. 
 
 To be concrete, consider the following integral which arises in the computation of the Compton scattering total cross section at next-to-leading order (NLO) in quantum electrodynamics \cite{comptoncrosssection}:
\begin{equation}
   I(s) = \int_1^s \frac{d s'}{\sqrt{(s'-1)(s'+3)}} \int_1^{s'} \frac{d s''}{s''}
   \label{eq:Compton_integral}
\end{equation}
One can often evaluate integrals like this by rationalizing the square-root and expanding in partial fractions. In this case, defining
$r=\frac{1}{2}(s+1+\sqrt{(s-1)(s+3)})$ the integral can be evaluated as
\begin{align}
    I(s)&=\int_1^{r(s)} d \ln r' \int_1^{r'} d \ln \frac{r''^2 - r'' +1}{r''}\\
    &=\int_1^{r(s)} \frac{dr'}{r'}\int_1^{r'} dr'' \left(-\frac{1}{r''}+\frac{1}{r''-z_1} + \frac{1}{r''-z_2} \right) \\
    &=\frac{2\pi^2}{9} - \frac{\ln^2 r}{2}  +(-\frac{i\pi}{3} + \ln r) \ln(r-z_1) + \ln(r z_1) \ln (r-z_2) \\
    &\hspace{15mm}- \Li_2(1-z_1) + \Li_2(1-r z_1) - \Li_2(1-z_2) + \Li_2(1-r z_2)
    \label{Is4}
\end{align}
where $z_1=\frac{1+i \sqrt{3}}{2}$ and $z_2=\frac{1-i \sqrt{3}}{2}$ are cube-roots of unity. Although this expression is correct, it can be simplified further to
\begin{equation}
    I(s) = -\frac{1}{3} \Li_2(-r^3)  + \Li_2(-r)- \frac{1}{2} \ln^2 r +\frac{\pi^2}{18} \label{IsLL}
\end{equation}
There is no existing software package (to our knowledge) which can  integrate and simplify such expressions automatically.  The goal of this paper is to explore how machine learning may assist in the simplification of expressions like $I(s)$ which arise from Feynman integrals. For this initial study, we restrict our considerations to classical polylogarithms and functions of a single variable where the problem is already  challenging. 

Simplifying expressions involving polylogarithms involves applying various identities. For example, the logarithm satisfies 
\begin{equation}
    \ln(xy) = \ln x + \ln y
\end{equation}
Identities involving the dilogarithm include reflection
\begin{equation}
\Li_2(x) =-\Li_2(1-x) + \frac{\pi^2}{6} - \ln(x)\ln(1-x) 
\, ,
\end{equation}
duplication
\begin{equation}
    \Li_2(x)= - \Li_2(-x) + \frac{1}{2} \Li_2(x^2)
    \, , \label{dup}
\end{equation}
and others. Similar identities are known for $\Li_3(x)$ and  higher weight classical polylogarithms. Although the complete set of identities is not known for general $\Li_n(x)$, known identities are often enough to simplify expressions that arise in physical problems. 

Even with a given set of identities, it is not always easy to simplify polylogarithmic expressions. There may be many terms, and it may take the application of multiple identities in a specific order for the expression to simplify. For example, applying the duplication identity in Eq.~\eqref{dup} to a single term will increase the number of dilogarithms, and it may be that only after applying reflection after duplication that terms will cancel. An analogy is a sacrifice in chess: initially it seems ill-advised, but after a subsequent move or two the sacrifice may be beneficial. The comparison with moves in a game suggests that perhaps reinforcement learning could be well-suited to the simplification of polylogarithms. We explore the reinforcement-learning approach in Section~\ref{sec:RL}. 

The simplification of expressions resulting from Feynman integrals is not of purely aesthetic interest. For example, constructing the function space \cite{Yan:2022cye} and understanding the singularities of scattering amplitudes \cite{Hannesdottir:2021kpd,Bourjaily:2020wvq} is essential to the $S$-matrix bootstrap program \cite{Eden:1966dnq}. However, singularities of individual terms in a Feynman integral may be absent in the full expression. Thus methods for simplifying iterated integrals have been of interest in physics for some time. A particularly effective tool is the {\bf symbol} \cite{polylogsmhv6l}. We review the symbol in Section~\ref{sec:symbol}. Briefly, for an iterated integral of the form
\begin{equation}
  \mathcal{I}_n(x) = \int d \ln R_1  \circ \cdots \circ d \ln R_n
\end{equation}
where the iterated integrals are defined such that
\begin{equation}
  \int_a^b d \ln R_1 \circ \cdots \circ d \ln R_n = \int_a^b \left(
  \int_a^t d \ln R_1  \circ \cdots \circ d \ln R_{n - 1}  \right)
  d \ln R_n (t)
  \label{eq:circdef}
\end{equation}
we have an associated symbol as
\begin{equation}
  \mathcal{S} [\mathcal{I}_n] = R_1 \otimes \cdots \otimes R_n \, .
\end{equation}
In particular,
\begin{equation}
  \mathcal{S} [\Li_n (x)] = - (1 - x) \otimes \underbrace{x \otimes
  \cdots \otimes x}_{n - 1}
  \label{Li2S}
\end{equation}
and
\begin{equation}
    \mathcal{S} \left[\frac{1}{n!}\ln^n x\right] = \underbrace{x \otimes
  \cdots \otimes x}_{n}
  \label{lnnS}
\end{equation}
The symbol is part of a broader algebraic operation called the coproduct \cite{duhrhopf}. It serves to extract the part of an iterated integral with maximal transcendental weight (e.g. $\cS[\Li_2(x) + \ln x + \pi] = \cS[\Li_2(x)]$).
The symbol has a product rule similar to the logarithm:
\begin{equation}
    \cdots\otimes f (x) g (y) \otimes \cdots = \cdots \otimes f (x) \otimes \cdots \quad +\quad  \cdots \otimes g (y) \otimes \cdots
\end{equation}
This product rule can in fact be used to generate all of the known dilogarithm identities (see Section \ref{sec:symbol}).

The symbol helps with the simplification of polylogarithmic functions. For example, the symbol of Eq.~\eqref{Is4} is
\begin{equation}
    \cS[I] = r\otimes \frac{r^2 - r +1}{r}\\
     =\frac{1}{3} [ r^3\otimes(r^3 +1)] - r \otimes (r+1) - r \otimes r
\end{equation}
which using Eqs.~\eqref{Li2S} and \eqref{lnnS} immediately gives the leading transcendentality part of Eq.~\eqref{IsLL}.
A more involved example is the simplification of two-loop 6-point function in super Yang-Mills theory from a 17-page expression to a few lines~\cite{polylogsmhv6l}. 

\begin{figure}
    \centering
    \begin{tikzpicture}
\node[block] (a) {Polylogarithmic expression\\[0.5cm]$\Li_2(\frac{1}{x})+\frac{1}{2} \ln^2(-x)$};
\node[block, right =2cm of a] (b){Simplified polylogarithmic expression \\[0.5cm] $\Li_2(x)$};
\node[block, below = 2cm of a] (c) {Symbol\\[0.5cm]$-[(1-\frac{1}{x})\otimes\frac{1}{x}] + [x\otimes x]$};
\node[block, below = 2.03cm of b] (d){Simplified Symbol \\[0.5cm] $(1-x)\otimes x$};
\draw[->, thick] ([yshift=-0.2cm]a.south) -- ([yshift=0.2cm]c.north);
\draw[->,line width=2.5, darkred] ([xshift=0.2cm]a.east) -- ([xshift=-0.2cm]b.west);
\draw[<-,line width=2.5, darkblue] ([yshift=-0.2cm]b.south) -- ([yshift=0.2cm]d.north);
\draw[->,thick] ([xshift=0.2cm]c.east) -- ([xshift=-0.2cm]d.west);
\node[above right = -0.5cm and 0.5cm of a, darkred] {RL};
\node[above right = 0.7cm and -1.5cm of d, darkblue] {Transformer};
\end{tikzpicture}
    \caption{Different approaches to simplifying polylogarithmic expressions. We can  try to learn the simplification directly, as we do with a reinforcement-learning algorithm  (top arrow). Or we can use a classical algorithm to compute the symbol and simplify the expression and then a transformer network to integrate the symbol (right arrow)}
    \label{fig:flow}
\end{figure}
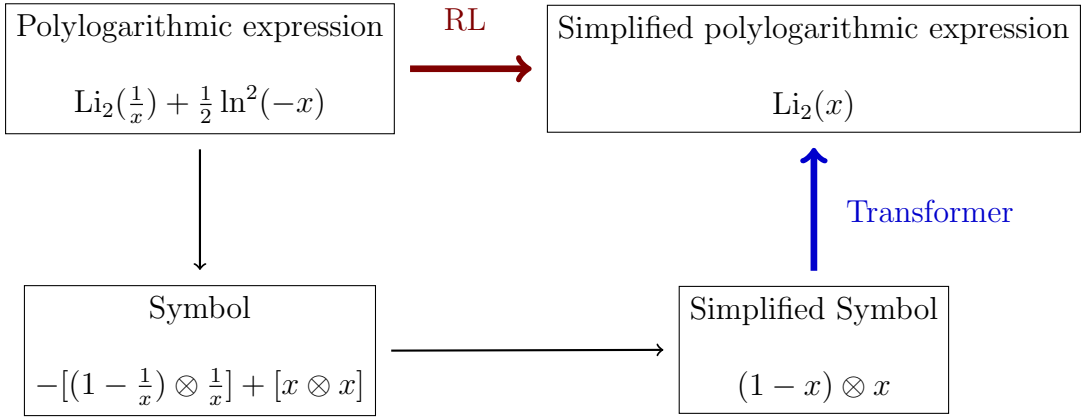

Unfortunately, the symbol does not trivialize the simplification of polylogarithmic functions. One can use the product rule to pull out factors of rational polynomial arguments, but getting the resulting symbol into sums of terms of the forms in Eqs.~\eqref{Li2S} and \eqref{lnnS} still requires intuition and often educated guesswork. While there are algorithms to compute the symbol from a given polylogarithm expression, there is not a known inverse algorithm, to compute the polylogarithm from the symbol. Machine learning, however, is good at inverse problems: one can view the integration of the symbol  as translation of natural language. Given a mapping from English (polylogarithms) to French (symbol), can we find the English translation (polylogarithmic form) of a given French phrase (known symbol)? Viewing the simplification problem as a sequence-to-sequence translation task allows us to explore a host of other machine learning techniques. In this paper, we will focus on using one of the most powerful known methods for natural language processing (NLP), transformer networks, to tackle the symbol-polylogarithm translation task. The way that reinforcement learning or transformer networks can be used in the simplification task is illustrated in Fig.~\ref{fig:flow}.

Machine learning has already proven useful for symbolic problems in mathematics. For instance, tree LSTM's were deployed to verify and complete equations \cite{arabshahi2018combining}, while transformers were shown to be able to solve ODE's, integrate functions \cite{fbtransformer} or solve simple mathematical problems \cite{DBLP:journals/corr/abs-1904-01557}. Finding new identities or checking the equivalence between two mathematical expressions falls into this category of symbolic tasks and has also been tackled using machine learning \cite{AllamanisCKS17, Zaremba}. In particular, reinforcement learning has provided interesting applications in symbolic domains, with contributions in theorem proving \cite{rltheorem,zombori2021towards, Wu2021TacticZeroLT, Lederman2020Learning}, symbolic regression \cite{petersen2021deep}
 or for the untangling of knots \cite{unknot,davies2021advancing}. Physics inspired applications of machine learning have emerged, predicting physical laws from data \cite{cranmer2020, tegmarkfeynmanai, guimera2020} or discovering its symmetries \cite{thaler2022, tegmark2022} but are for the most part focused on problems with numerical input data. Using machine learning in purely symbolic mathematical environments is often quite challenging: when solving a given problem the solution has to be exact and the path to reach it is typically narrow. For example when simplifying equations we have to take a sequence of identities in a precise order where any deviation from it leads to a completely different expression. Those challenges are discussed in \cite{contrastiveRLPoesia} where the use of contrastive learning \cite{aaron2019, Haotian2020} was key in solving various symbolic domains. 

We begin in Section~\ref{sec:math} with a review of the relevant mathematics of polylogarithms and symbols that we use later in the paper. Section \ref{sec:classical_algos} describes our ``classical" approach to simplifying polylogarithm expressions, while Section~\ref{sec:RL} presents our reinforcement learning based technique. Section~\ref{sec:transformer} explores the transformer network's perspective in the translation task between a symbol and a polylogarithmic expression. 
A example complete non-trivial application of the transformer network is given in Section~\ref{sec:summary}.
Some conclusions and the outlook is discussed in Section~\ref{sec:conc}.

\section{Mathematical preliminaries~\label{sec:math}}
In this section we briefly review the classical polylogarithms and the symbol map. The reader familiar with the mathematics of polylogarithms can safely skip this section. More details can be found in~\cite{duhrhopf, dilogkirilov}.

\subsection{Classical polylogarithms}
The classical polylogarithm $\Li_n(x)$ is defined as an iterated integral, most conveniently expressed recursively as
 \begin{equation}
     \Li_n(x) = \int_0^x dt \, \frac{\Li_{n-1}(t)}{t}
 \end{equation}
with the base case
 \begin{equation}
     \Li_1(x) \equiv -\ln(1-x) = \int_0^x \frac{d t}{1-t}
 \end{equation}
 so that
 \begin{equation}
     d \Li_n(x) = \Li_{n-1} (x) \frac{dx }{x} \, .
 \end{equation}
In general, one has to be careful about the branch cuts of polylogarithmic functions. $\ln(x)$ has a branch point at $x=0$ and is conventionally defined to have a branch cut on the real axis for $x<0$. $\Li_n(x)$ for $n\ge 1$ has a branch point at $x=1$ with a branch cut from $1<x<\infty$  on the real axis and a branch point at $x=0$ on a higher Riemann sheet. For numerical checks, it is important to keep track of the branch of the polylogarithm, but for symbolic manipulation we can largely ignore the subtleties of the analytic structure.

The logarithm satisfies the product rule
\begin{equation}
    \ln(x y) = \ln(x) + \ln (y)
\end{equation}
 which is the only identity needed to simplify logarithms.
 
The dilogarithm satisfies a number of identities, such as
    \begin{align}
(\textit{inversion}) \quad \Li_2(x)&=-\Li_2\left(\frac{1}{x}\right) - \frac{\pi^2}{6} - \frac{\ln^2(-x)}{2}\label{inversion}\\[0.8em]
(\textit{reflection}) \quad \Li_2(x)&=-\Li_2(1-x) + \frac{\pi^2}{6} - \ln(x)\ln(1-x)\label{reflection}\\[0.8em]
(\textit{duplication}) \quad \Li_2(x)&= - \Li_2(-x) + \frac{1}{2} \Li_2(x^2)\label{duplication} 
\end{align}
These identities can be derived by taking derivatives, using the logarithm product rule and then integrating. 
Inversion and reflection form a closed group\footnote{This is often called the Coxeter group with the presentation $\langle R,I \, |\, (IR)^3=1, I^2=1, R^2=1\rangle$.}, in that applying them iteratively generates only 6 different arguments for the dilogarithm
\begin{equation}
\label{sixterms}
    \Li_2(x),\quad \Li_2(1-x), \quad \Li_2\left(\frac{1}{x}\right),
    \quad \Li_2\left(\frac{x}{1-x}\right),
    \quad \Li_2\left(\frac{x-1}{x}\right),
    \quad \Li_2\left(\frac{1}{1-x}\right),    
\end{equation}
Thus without something more complicated like duplication, the simplification of an expression involving dilogarithms would be a finite problem. The duplication identity is part of a broader set of identities
\begin{equation}
    \Li_2(x) = n \sum_{z^n=x} \Li_2(z), \quad n=1,2,3,\cdots
\end{equation}
For the dilogarithm, the five term identity:
\begin{multline}
    \Li_2(x) + \Li_2(y) + \Li_2\left(\frac{1-x}{1-xy}\right) + \Li_2(1-xy) + \Li_2\left(\frac{1-y}{1-xy}\right) \\*
    = \frac{\pi^2}{2} - \ln(x)\ln(1-x)  - \ln(y)\ln(1-y) - \ln\left(\frac{1-x}{1-xy}\right) \ln\left(\frac{1-y}{1-xy}\right)
\end{multline}
is a kind of master identity~\cite{wojtkowiak1996functional, zagierdilog}
from which other identities can be derived by judicious choices of $y(x)$. 
For instance taking $y=x$ gives
\begin{equation}
    \Li_2(x) + \frac{1}{2}\Li_2(1-x^2) + \Li_2\left(\frac{1}{1+x}\right)+\ln(x)\ln(1-x)+\frac{1}{2}\ln^2\left(\frac{1}{1+x}\right)  = \frac{\pi^2}{4}
\end{equation}
which generates the duplication identity after using the reflection and inversion on the second and third dilogarithms. In fact all dilogarithm identities involving a single variable are expected to be recovered from the 5 term identity in this way \cite{dilogkirilov}.

Higher-weight polylogarithms in $n$ satisfy similar identities, such as
\begin{equation}
    \Li_3(x) = \Li_3\left(\frac{1}{x}\right) - \frac{1}{6} \ln^3 (-x) - \frac{\pi^2}{6}\ln(-x)
\end{equation}
and so on. It should be noted however that fewer functional identities are known for polylogarithms of higher weight.

\subsection{Multiple polylogarithms and the symbol map}\label{sec:symbol}
A generalization of classical polylogarithms are the multiple or Gonchorov polylogarithms \cite{goncharov1mpl} defined iteratively as
\begin{equation}
  G (a_1, \cdots, a_n ; x) = \int_0^x \frac{d t}{t - a_1}\,  G (a_2, \cdots, a_n ; t)
  \label{Gdef}
\end{equation}
starting from
\begin{equation}
    G(a;x)\equiv \ln (x-a)
\end{equation}
and with 
\begin{equation}
    G(\underbrace{0, \cdots 0}_{n}; x) \equiv \frac{1}{n!} \ln^n(x) \, .
\end{equation}
We can write these functions in an alternative notation as
\begin{equation}
      G (a_1, \cdots, a_n ; x) =\int_0^x d \ln (x-a_1) \circ \cdots \circ d \ln (x-a_n)
\end{equation}
which follows the definition in Eq.~\eqref{eq:circdef}. The multiple polylogarithms form a Hopf algebra, which is an algebra with a product and a coproduct~\cite{duhrhopf}; this algebra also has a shuffle product \cite{ree1958lie}.
The details of the algebraic structure are not of interest to us at the moment. All we need to know is that the maximum iteration of the coproduct, called the {\bf symbol} \cite{polylogsmhv6l} and denoted $\cS$ extracts the logarithmic differentials from an iterated integral
\begin{equation}
    \cS\left[\int_a^b d \ln R_1  \circ \cdots \circ d \ln R_n \right]
    =R_1 \otimes \cdots \otimes R_n
\end{equation}
so that
\begin{equation}\label{symbolpolylogs}
  \cS\big[G(a_1,\ldots,a_n;x)\big] = (x-a_1) \otimes \cdots \otimes (x-a_n)
\end{equation}
with the special cases
\begin{equation}\label{symbolpolylogs2}
  \cS\big[\Li_n (x)\big] = - (1 - x) \otimes \underbrace{x \otimes
  \cdots \otimes x}_{n - 1}
\end{equation}
and
\begin{equation}
    \mathcal{S} \left[\frac{1}{n!}\ln^n x\right] = \underbrace{x \otimes
  \cdots \otimes x}_{n}
  \label{lnnS2}
\end{equation}
The symbol acting on a complex number vanishes, e.g.  $\cS[c]=0$. So a function cannot always be exactly reconstructed from its symbol. One can, however, reconstruct the leading-transcendentality part of a function.\footnote{Transcendentality is defined so that $\Li_n(x)$ and $\ln^n(x)$ have transcendentality $n$. The transendentality of a product of two functions is the sum of the functions' transcendentality. E.g. $\ln^n(x) \Li_m(x)$ has transcendentality $n+m$. So $\Li_3(x)$ and $\Li_3(x) + 3 \Li_2(x)$ have the same symbol.}

The symbol is powerful because it is  additive
\begin{equation}
  \mathcal{S} \big[f (x) + g (y)\big] = S\big[f (x)) + S (g (y)\big]
\end{equation}
and satisfies the product rule
\begin{equation} \label{productrule}
  \cdots\otimes f (x) g (y) \otimes \cdots = \cdots \otimes f (x) \otimes \cdots +
  \cdots \otimes g (y) \otimes \cdots
\end{equation}
Similar to how $\cS[c]=0$, at the level of the symbol, any multiplicative constant $c$ can also be discarded:
\begin{equation}
  \cdots\otimes \, c f (x) \otimes \cdots = \cdots \otimes f (x) \otimes \cdots
\end{equation}
This behavior is consistent with the symbol only capturing the highest
transcendental weight portion of a given function.

As an example of the use of a symbol, consider the reflection dilogarithm identity. Taking the symbol we find
\begin{equation}
\cS\left[\Li_2(x)+\Li_2\left(\frac{1}{x}\right)+\frac{1}{2}\ln^2 (-x)\right] 
=- [(1-x)\otimes x] - [(1-\frac{1}{x}) \otimes\frac{1}{x} ]+ [ x\otimes x]  =0
\end{equation}
where the product rule was used in the last step. 

For transcendental-weight-two functions (dilogarithms), the symbol has two entries. It can be helpful to decompose such symbols into symmetric and antisymmetric parts. The symmetric part can always be integrated into products of logarithms
\begin{equation}
    \cS\big[ \ln x \ln y\big] = x \otimes y + y \otimes x
\end{equation}
The anti-symmetric part is harder to integrate. It can be convenient to express dilogarithms in terms of the the Rogers $L$ function
\begin{equation}
    L(x) \equiv \Li_2(x) + \frac{1}{2} \ln x \ln (1-x)
\end{equation}
whose symbol is anti-symmetric
\begin{equation}
    \cS\big[L(x)\big] = \frac{1}{2} x \otimes(1-x)-\frac{1}{2} (1-x) \otimes x
\end{equation}
So the antisymmetric part of a symbol can be integrated if one can massage it into a difference of terms of the form $x\otimes(1-x)$. For higher-weight functions, one can antisymmetrize the first two entries of their symbols, but there is not a canonical way to write or integrate these symbols.

For a neat application, define $\rho = \frac{\sqrt{5}-1}{2}$ so that $\rho^2+\rho-1=0$ and hence
\begin{equation}
    \rho^3 - 1 = \rho(1-\rho)-1 = -(\rho^2-\rho+1)=-\frac{\rho^3+1}{\rho+1}
    \label{rhoid} \, .
\end{equation}
Now consider the combination~\cite{dilogkirilov}.
\begin{equation}
    F(\rho) = L(\rho^6) - 4 L(\rho^3) -3 L(\rho^2)+ L(\rho) \label{Frel} 
\end{equation}
The symbol of $F(\rho)$ is the antisymmetrized version of
\begin{align}
\cS[F(\rho)] &=\rho^6 \otimes (1-\rho^6) - 4[\rho^3 \otimes(1-\rho^3)] - 3[\rho^2\otimes(1-\rho^2)] +6 [\rho \otimes(1-\rho)] \\*
    &=\rho \otimes(1-\rho^6)^6 - \rho\otimes(1-\rho^3)^{12} - \rho \otimes(1-\rho^2)^6 + \rho \otimes(1-\rho)^6\\*
    &= \rho \otimes \frac{(\rho^3+1)^6}{(\rho^3-1)^6(\rho+1)^6}
\end{align}
Using Eq.~\eqref{rhoid}, this simplifies to $\cS[F(\rho)] = \rho \otimes 1 =0$. We conclude that the combination in Eq.~\eqref{Frel} is free of dilogs. Indeed, $F(\rho)=\frac{7\pi^2}{30}$ as computed by Coxeter in 1935 using a geometric argument~\cite{coxeter1935}, and in~\cite{dilogkirilov} using the 5-term dilogaritm identity. 
Using the symbol vastly simplifies the derivation of Coxeter's formula.

A powerful application of the symbol is to simplify polylogarithmic expressions through algebraic manipulations. 
However, after simplifying the symbol, it is not always easy to integrate it back into a function. 
As an example, a Feynman integral might lead us to consider
\begin{equation}
  \mathcal{I} (x) = \int_0^x d y_1  \frac{1}{y_1} \int_0^{y_1} d y_2 
  \frac{1}{1 - y_2} \int_1^{y_2} d y_3  \frac{1}{y_3}
  \label{IIform}
\end{equation}
Doing the integral in $\texttt{Mathematica}$ with $0<\operatorname{Re}(x)<1$ and $\operatorname{Im}(x)<0$ gives
\begin{multline}
  \mathcal{I} (x) = - 2 \text{Li}_3 (x) + i \pi \ln(1-x) \ln(x)\\
  +\frac{1}{6} \ln (x)  \left( - 6 \text{Li}_2 \left(
  \frac{1}{1 - x} \right) + 12 \text{Li}_2 (x) - 3 \ln^2 (1 - x) + 6 \ln (x) \ln (1 - x)  + \pi^2 \right)  .
\end{multline}
The symbol of this function is easiest to read directly from the integral form
in Eq.~\eqref{IIform}
\begin{align}
  \cS[\mathcal{I} (x)] &= -  x \otimes (1 - x) \otimes x\\
   &= - 2(1-x) \otimes x \otimes x +2 (1-x) \otimes x \otimes x - x \otimes (1-x) \otimes x\\
   &= \cS[- 2 \text{Li}_3 (x) + \text{Li}_2 (x) \ln (x)]
\end{align}
The second line required massaging the symbol so that the symbol of $\Li_3(x)$ was manifest and then recognizing the remainder as the symbol of a simpler function. The result is that 
\begin{equation}
  \mathcal{I}(x) = - 2 \text{Li}_3 (x) + \text{Li}_2 (x) \ln (x)
\end{equation}
up to lower transcendental-weight functions (there are none in this case).

As this example indicates, it is often possible to simplify a function using the symbol, but such a simplification requires some intuition. Simplicity and intuition are not easy to encode in classical algorithms, however, they are traits that machine learning has been known to harness. In Section \ref{sec:transformer} we will see a concrete application by deploying transformer networks to integrate the symbol.

\section{Classical algorithms}\label{sec:classical_algos}

\noindent
Before introducing  Machine Learning (ML) algorithms, we briefly summarize how one could simplify polylogarithms using standard algorithmic approaches. This will serve as a good benchmark for any ML implementation. We start by discussing algorithms for reducing polylogarithm expressions using identities. If we restrict the identities needed for a specific problem to a finite set (such as inversion, reflection and duplication), then it is not difficult to write a decent algorithm. We will see however that the main disadvantage of such methods lies in their computational cost scaling.


\subsection{Exploring the simplification tree} \label{sec:mbfs}

\noindent
\textbf{Breadth-first search} \quad  Starting from a given expression, such as $F(x) = \Li_2{(x)}+ \Li_2\left(\frac{1}{1-x}\right)$, one approach to find a simplified form is a breadth-first search. We construct a {\it{simplification tree}}, by applying iteratively all possible identities to all distinct dilogarithm terms. Each new expression obtained in such manner is a node of the simplification tree. In this way we can exhaustively explore the set of all possible expressions reached by our identities up to a given depth. The depth of this tree will correspond to the number of moves required to reach our starting point, the root of the tree. If the number of terms in a given expression always remained 2, the scaling for the number of nodes would go as $6^N$ at depth $N$.  The scaling is made worse by the presence of the duplication identity, which can increase the number of distinct dilogarithm terms. Clearly this kind of exhaustive search is extremely inefficient.  Nonetheless, provided we have enough computational resources we can fully explore the tree and we are assured that any solution we find will be the one requiring the fewest identities to reach. In practical applications where we can have up to $\mathcal{O}(10^3)$  dilogarithm terms \cite{Bonciani2011} this is not reasonable.

\bigskip
\noindent
\textbf{Modified Best-first search}  \quad Another way of searching the simplification tree is by following a best-first search. At each given iteration we once again apply all identities to all distinct dilogarithm terms. This time however we choose only one node as our next starting point, the one corresponding to the simplest expression. We define this ``best" node as the one with the fewest dilogarithms. In the event of a tie (for instance when no obvious simplifications are achieved by using a single identity) we retain the node that corresponds to applying a reflection identity on the first dilogarithm term. After doing so, we apply a cyclic permutation to the dilogarithm terms of the expression, preventing the algorithm from getting stuck in a loop.
This algorithm is expected to perform well if only reflection $R$ and inversion $I$ are considered. Indeed, these identities satisfy 
\begin{align}
    R^2 f(x) = f(x); \quad I^2 f(x) =f(x); \quad (I R)^3 f(x) =f(x)
\end{align}
generating a closed group of order 6 as in Eq.~\eqref{sixterms}. Allowing the algorithm to take a reflection identity on each term sequentially guarantees that any two matching terms will be identified by our best first search procedure, using one inversion at most. For instance we could have the following simplification chain
\begin{equation}
    RI f(x) - f(x) \xrightarrow{\text{tie}} R^2 I f(x) - f(x) \xrightarrow{R^2=1\, \& \,\text{cyclic}}  -f(x)  + I f(x) \xrightarrow{\text{best}} 0 \,.
\end{equation}

\noindent
The number of evaluations needed at each depth of the search tree is given by $N_{\text{eval}} =N_{\text{ids}} \, N_{\text{terms}} $. This scaling can also be prohibitive for computations of physical interest. Guiding the search tree would provide a crucial improvement and we will discuss in Section \ref{sec:RL} how Reinforcement Learning techniques could be of some use.

\subsection{Symbol Approach}\label{sec:symbol-integration}
We next consider a classical algorithm to construct a polylogarithmic expression given its symbol, i.e. to ``integrate the symbol", to benchmark our ML implementation.
As mentioned in Section~\ref{sec:symbol}, the symbol assists in the simplification of polylogarithmic expressions by reducing the set of identities to the product rule, so simplification can be done with standard polynomial-factorization algorithms. For example, the \texttt{mathematica} package \texttt{PolyLogTools}~\cite{polylogtools} can automate the computation of the symbol of a given polylogarithmic expression and its simplification. 
The tradeoff is that it is not so easy to integrate the simplified symbol back into polylogarithmic form. 

In general, symbols can be extraordinarily complicated. To have a well-defined problem and to guarantee physical relevance, we consider as example the symbol of the various transcendentality-two terms in the Compton scattering total cross section at NLO~\cite{comptoncrosssection}. 
The original function space is composed of 12 transcendental weight-two functions $g_{1,\cdots 12}$, which each contain complex numbers and irrational numbers like $\sqrt{3}$ and $\sqrt{5}$. The corresponding symbols after mild simplification are:
\begin{align}
    \label{eq:Compton_function_symbol}
    \cS[g_1]&=(1-y^2)\otimes (1+y), \quad 
    \cS[g_2]=y\otimes (1-y^2), \quad 
    \cS[g_3]=(1+3y^2)\otimes (1-y^2)\notag \\
    \cS[g_4]&=\frac{1-y^2}{1+3y^2}\otimes y,\quad \cS[g_5]=\frac{1-y^2}{1+3y^2}\otimes (1+y^2),\quad \cS[g_6]=(1-y^2)\otimes (1-y)\notag\\
    \cS[g_7]&=(1+3y^2)\otimes \frac{1+y}{1-y},\quad \cS[g_8]=(1-y^2)\otimes (1+3y^2),\quad 
    \cS[g_9]=y\otimes (1+3y^2)\notag\\
    \cS[g_{10}]&=(1+3y^2)\otimes (1+3y^2),\quad \cS[g_{11}]=\frac{1-y^2}{y^2}\otimes (1-5y^2)\notag\\
    \cS[g_{12}]&=y\otimes \frac{(1+3y^2)^2}{(1-y^2)^2}+(1+2y^2-3y^4)\otimes \sqrt{\frac{1-y^2}{1+3y^2}}
\end{align}
Note that these symbols are purely real. Also note that the polynomials appearing in the rational function arguments have degree at most 4 and integer coefficients no larger than 5. Thus it will not be unreasonable to consider the simpler problem of simplifying expressions with restrictions on the set of rational functions appearing as arguments of the polylogarithms.

The general strategy for integration is as follows :
\begin{itemize}
    \item Integrate \textit{symmetric} terms that give rise to logarithms: 
    \begin{equation}
    (y+a)\otimes (y+b)+(y+b)\otimes (y+a)\quad \longrightarrow \quad \ln(y+a)\ln(y+b)
    \end{equation}
    \item Integrate terms with \textit{uniform powers} $(a+by^n)\otimes (c+dy^n)$ into polylogrithms. The idea is to find two constants $y_1$ and $y_2$ such that 
    \begin{equation}
    y_1 (a+by^n)\otimes y_2 (c+dy^n)=(1-g(y))\otimes g(y)\quad \longrightarrow \quad -\Li_2 [g(x)]
    \end{equation}
    If $ad-bc\neq 0$, solving this equation gives $y_1=\frac{d}{ad-bc}$, $y_2=\frac{b}{bc-ad}$, which guarantees the integrability of the uniform power terms. For terms like $(a+by^n)\otimes y$, we can apply the product rule: $\frac{1}{n}(a+by^n)\otimes y^n$ and integrate it similarly.
    \item Search for terms that can be combined to $(a+by^n)\otimes (c+dy^n)$. For example, we search for terms like
    \begin{align}
        (a+by^2)\otimes (c+y)+(a+by^2)\otimes (c-y)&=(a+by^2)\otimes (c^2-y^2)\notag\\
        (a+by^3)\otimes (c-y)+(a+by^3)\otimes (c^2+cy+y^2)&=(a+by^3)\otimes (c^3-y^3)\notag\\
        \cdots\cdots\qquad\qquad &
    \end{align}
    which can be integrated following step 2. For the remaining symbol, we can try feeding terms like $1\pm y$. Explicitly,
    \begin{equation}
        y\otimes(1+y+\cdots y^{2m})=y\otimes(1-y^{2m+1})-y\otimes (1-y)
    \end{equation}
    and both terms can be integrated directly.
\end{itemize}

\noindent
Under this algorithm, we integrate all basis symbols successfully except $g_7$, and all expressions are free of square roots and complex numbers. For example we find
\begin{equation}
    \cS[g_{11}]=\cS\left[\Li_2(1-5y^2)-\Li_2\left(\frac{5y^2-1}{4}\right)\right] \, .
\end{equation}
The remaining basis $g_7$ requires introducing another variable $z=\frac{1-y}{1+y}$, and our algorithm gives
\begin{equation}
    \cS[g_7]=2(1+z)\otimes z-(1-z+z^2)\otimes z=\cS\left[-3\Li_2(-z)+\frac{1}{3}\Li_2(-z^3)\right] 
\end{equation}

In conclusion, while it is often possible to integrate the symbol with a classical algorithm, one must proceed on a case-by-case basis. That is, there often seem to be nearly as many special cases as expressions, even at weight two. As we will see in Section \ref{sec:transformer}, transformer networks seem to be able to resolve many challenging weight-two expressions in a more generic way. 

\section{Reinforcement Learning \label{sec:RL}}
 As detailed in previous sections our goal is to simplify expressions involving polylogarithms. 
Sometimes it is possible to simplify polylogarithmic expressions by hand, by choosing both an identity and the term to apply it on. The difficulty in this procedure lies in figuring out which choice of identity will ultimately lead us to a simpler expression, maybe many manipulations down the line. In practice this often relies on educated guesswork, requiring experience and intuition.
Such a problem seems particularly well adapted for Reinforcement Learning (RL). RL is a branch of machine learning where an \textit{agent} interacts with an \textit{environment}. Based on a given \textit{state} of the environment the agent will decide to take an \textit{action} in order to maximize its internal \textit{reward function}. Our goal is to explore whether RL can be useful in reducing linear combinations of polylogarithms. 

We will consider a simplified version of this general problem by restricting ourselves to dilogarithmic expressions which can be reduced using the actions of inversion, reflection and duplication shown in Eqs.~\eqref{inversion}, \eqref{reflection} and~\eqref{duplication}. 
Since simplifying single logarithms and constants is not difficult, we drop these terms from our calculus (since we can keep track of each action taken we are able to add back the discarded constants and logarithms at the end). So all of our simplifications will be made at the level of dilogarithms only.

It may seem that we have oversimplified the problem to the point where it is no longer of interest in physical applications, like computing Feynman integrals. However, if one cannot solve this simple problem, then there is no hope of solving the more complex one of physical interest. We view our toy dilogarithm problem analogously to the problem of integrating relatively simple functions of a single variable in~\cite{fbtransformer} or the factorization of polynomials like $x^2+3x-10$ tackled in~\cite{DBLP:journals/corr/abs-1904-01557}.

\subsection{Description of the RL agent} \label{sec:rl-desc}
We have motivated the following  problem:

\bigskip
\noindent
\textbf{Problem statement} \quad What is the simplest form of a linear combination of dilogarithms with rational functions as arguments?  We consider only forms that can be simplified using the inversion, reflection and duplication identities. By using these identities to generate more complicated forms from simple ones, we can guarantee that all the examples in our data set simplify.

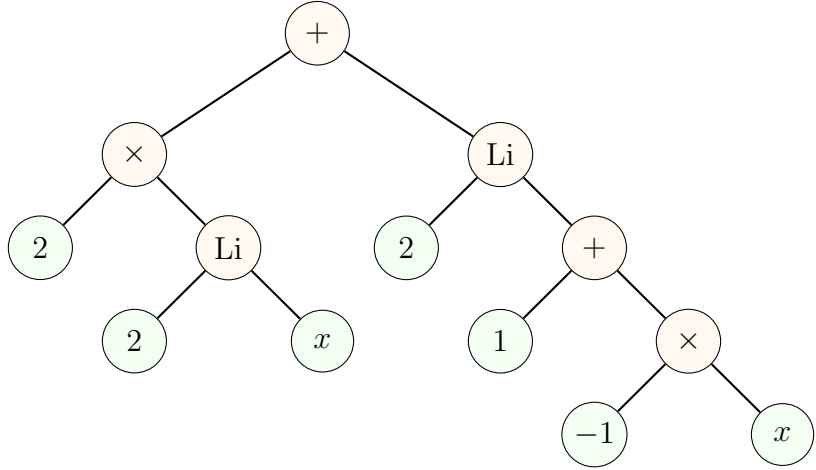
\begin{figure}[t]
\centering
    \begin{tikzpicture}[node distance=1.75cm]
    \node (op1) [operator] {$+$};
    \node (op2) [operator, below left =1cm and 1.8cm of op1] {$\times$};
    \node (leaf1) [leaf, below left of=op2] {$2$};
    \node (op3) [operator, below right of=op2] {$\Li$};
    \node (leaf2) [leaf, below left of=op3] {$2$};
    \node (leaf3) [leaf, below right of=op3] {$x$};
    \node (op4) [operator, below right=1cm and 1.8cm of op1] {$\Li$};
    \node (leaf4) [leaf, below left of=op4] {$2$};
    \node (op5) [operator, below right of=op4] {$+$};
    \node (leaf5) [leaf, below left of=op5] {$1$};
    \node (op6) [operator, below right of=op5] {$\times$};
    \node (leaf6) [leaf, below left of=op6] {$-1$};
    \node (leaf7) [leaf, below right of=op6] {$x$};
    
\draw [line] (op1) -- (op2);
\draw [line] (op2) -- (leaf1);
\draw [line] (op2) -- (op3);
\draw [line] (op3) -- (leaf2);
\draw [line] (op3) -- (leaf3);
\draw [line] (op1) -- (op4);
\draw [line] (op4) -- (leaf4);
\draw [line] (op4) -- (op5);
\draw [line] (op5) -- (leaf5);
\draw [line] (op5) -- (op6);
\draw [line] (op6) -- (leaf6);
\draw [line] (op6) -- (leaf7);

    \end{tikzpicture}
    \caption{Tree structure of $2\, \Li_2(x) + \Li_2(1-x)$.}
    \label{treeexpr1}
\end{figure}

\bigskip
\noindent
\textbf{State space} \quad The state space corresponds to the set of linear combinations of dilogarithms whose arguments are rational functions over the integers. This state space is preserved under inversion, reflection and duplication. We encode the state space so that the agent is confronted with numerical data as a opposed to symbolic data. For this purpose we first translate any given equation into prefix notation, taking advantage of the fact that any mathematical equation can be represented by a tree-like structure. For instance the mathematical expression $2\, \Li_2(x)+\Li_2(1-x)$ is parsed as [`add', `mul', `+', `2', `polylog', `+', `2', `x', `polylog', `+', `2', `add', `+', `1', `mul', `-', `1', `x'] which is represented by the tree of Fig.~\ref{treeexpr1}. 

\smallskip
\noindent
For the constants  we use a numeral decomposition scheme, where integers are explicitly written as $\sum_j c_j \times 10^j$. This implies adding the extra word `10' to our vocabulary compared to \cite{fbtransformer}, but typically leads to a better performance, following the observations made in \cite{graphmr}. In total we have $n_\text{words} = 20$ words.

\medskip
\noindent
\textbf{Word embedding}  \quad Our prefix expression is passed through a word embedding layer $E_W$ for which different choices are considered. Since we have a small number $n_\text{words}$ of words (`2', `polylog', `+', etc.)  that are required to parse our mathematical expression we experimented with either label encoding, one-hot encoding \cite{Pargent2022} or a dedicated embedding layer at the word level . We found that one-hot encoding provided the most stable results and it will be our encoding of choice.  The mapping between the words of the dictionary $D$ and the target vector space is given by $E_W : D \rightarrow \mathbb{R}^{n_{\text{words}}}$. Each word will correspond to a distinct unit vector in this space.

\medskip
\noindent
\textbf{Sentence embedding}  \quad We also need to specify an encoding scheme for the sentence as a whole. The simplest choice is to consider a finite length for our input, $L_{\text{max}}$ and to pad the observation vector. The padding value is a vector of $0$'s following our one-hot encoding implementation. The resulting one-hot encoded equation has dimensions $L_{\text{max}} \times n_\text{words}$. Any equation with a prefix notation expression longer than $L_{\text{max}}$ will be deemed invalid and will be discarded.

We also consider Graph Neural Networks (GNN) for encoding the prefix expression (sentence). The graph is given by the tree structure of the mathematical expression, where the nodes take as value the one-hot encoded word. Our architecture is implemented using \texttt{PyTorch Geometric} \cite{torch-geometric} and consists of multiple message passing layers using a mean aggregation scheme along with a final global pooling layer across the nodes. In that way the entire mathematical expression can be represented as a vector of size $n_\text{emb}$, where $n_\text{emb}$ is the embedding dimension used by the message passing layers. 

\bigskip
\noindent
\textbf{Action space}\quad Each identity (reflection, inversion or duplication) corresponds to one of the actions that the agent can take. We also need to specify the dilogarithm term that we are acting on. For this purpose we take the convention that an identity is always applied on the first term of the linear combination. To make sure that the agent has access to all terms we implement a fourth action: the possibility of doing a cyclic permutation through the dilogarithm terms in the expression. The set of actions is illustrated in the following example:

\begin{equation}
2 \Li_2{(x)} + \Li_2{(1-x)} \rightarrow
\left\{\begin{array}{cc}
  -  \Li_2{(1-x)}   & (\textit{reflection}) \\[0.5em]
   -  2\Li_2{(1/x)} + \Li_2{(1-x)}  & (\textit{inversion})  \\[0.5em]
    -  2 \Li_2{(-x)} + \Li_2{(x^2)} + \Li_2{(1-x)}  &  (\textit{duplication}) \\[0.5em]
 \Li_2{(1-x)} +    2 \Li_2{(x)}  & ( \textit{cyclic permutation})
\end{array}\right.
\end{equation}

\smallskip
\noindent
After taking an action we process the expressions to combine relevant terms whenever possible. Following this convention, in our example the result of the reflection identity is directly $-  \Li_2{(1-x)}$. Simplifying the expressions after each action allows us to keep a small action space. Since the first dilogarithm term has now more importance, being the term that we are acting on, for the GNN embedding we extend the state vector by adding to it the embedding corresponding to the first dilogarithm term only. 

\bigskip
\noindent
\textbf{Reward function}\quad Our goal is to simplify linear combinations of dilogarithms. For that purpose a natural reward function is one that penalizes long sequences. In particular at each time step $t$ we could consider imposing a penalty based on the expression's length: 
\begin{equation}
\tilde{r}_t = -\lambda_1 \, N^\text{words}_t - \lambda_2 \, N^\text{dilogs}_t   \, , 
\end{equation}
where $N^\text{words}_t$ is the length of the sequence in prefix notation, $N^\text{dilogs}_t$ the number of distinct dilogarithm terms and $\lambda_1,\lambda_2$ a set of parameters. We observed that such a reward function is very sensitive to the choices of $\lambda_1$ and $\lambda_2$. It also leads to the duplication action being mostly ruled out by the agent, as any ``wrong" application leads to an immediate increase in the length of the expression.

\smallskip
\noindent We found a better reward function to be
\begin{equation} \label{eq:rew1} 
    r_t = \left\{
\begin{array}{cl}
1 & \text{if } N^\text{dilogs}_t< N^\text{dilogs}_{t'} \ \ \forall \ t' < t\\
0 & \text{ else} 
\end{array}
\right. \,.
\end{equation}

\noindent
 Choosing this reward function allows our agent to explore a larger portion of the state space since no penalty is imposed on the length of the sequence. 

Adopting the second reward scheme, one could be concerned with the possibility of having cyclic trajectories not being penalized by the agent. For instance, successive applications of the reflection action lead to recurrences
\begin{equation}
    \Li_2{(f(x))} \xrightarrow{\text{reflection}} \Li_2{(1-f(x))} \xrightarrow{\text{reflection}} \Li_2{(f(x))} \,.
\end{equation}

\noindent
To dissuade the agent from following such behavior we can add an extra penalty term
\begin{equation} \label{eq:rew2}
    \Delta r_t = \left\{
\begin{array}{cl}
-\lambda_r & \text{if } a_{t-1} = a_t \text{ and } N^\text{dilogs}_{t-1} \leq N^\text{dilogs}_{t}\\
0 & \text{ else} 
\end{array}
\right.
\end{equation}
where $a_t$ is the action taken at the time step $t$ and $N^\text{dilogs}_{t}$ the corresponding number of dilogarithm terms. This penalty term checks whether we are repeating actions, without shortening the expression length.
Adding this constraint should guide us away from cyclic trajectories.

Since RL is used to solve Markov Decision Processes we must ensure that our environment satisfies the Markov property, namely that any future state reached depends only on the current state and on the action taken. To respect this property and get accurate expected rewards we extend our state vector (after it has passed through the sentence embedding layer) by adding to it $\left[a_{t-1}, N_{t-1}^{\text{dilogs}}, \displaystyle \min_{t' < t} N_{t'}^{\text{dilogs}} \right]$. This additional information can now be used to estimate the full reward $r_t + \Delta r_t$.

\bigskip
\noindent
\textbf{RL Agent}\quad For the RL agent we tried Proximal Policy Optimization (PPO) \cite{ppo} and Trust Region Policy Optimization (TRPO) \cite{trpo}, both implemented within \texttt{stable-baselines3} \cite{stable-baselines3}. These agents are on-policy algorithms, where the exploration phase is governed by using actions sampled from the latest available version of the policy. The trajectories collected (actions taken, states, rewards) during the evolution of the agent are used to update the policy by maximizing a surrogate advantage\footnote{We note that \texttt{stable-baselines3} relies on Generalized Advantage Estimation (GAE) \cite{gae} when computing the policy gradient. In practice this is used to provide a trade-off between using real rewards (small bias/ high variance) and estimations from the value network (high bias/ low variance).}. We refer the reader to the aforementioned references for an in-depth explanation. 

In our experiments we observed that TRPO gave the more consistent and stable training. The particularity of the TRPO algorithm lies in that the policy updates are constrained, making the learning process more stable. By estimating the KL-divergence between the old and the new policy one takes the largest step in parameter space that ensures both policies' action distributions remain close. The TRPO algorithm  can also guarantee a monotonic improvement of the policy during updates, with increasing expected returns, although at the cost of longer training times.

\bigskip
\noindent
\textbf{Network Architecture}\quad The TRPO agent has a policy $\pi_\theta(a|s)$ and a value function $V_\phi(s)$, both parameterized by neural networks, for which we design a custom architecture. The input for both networks will be an observation $s$, which is the output of the sentence embedding layer, where we add an extra flattening layer if required. The final output is a scalar for the value net and a vector of dimensions $n_\text{actions}$ for the policy net. We opt for one shared layer of size 256 between the policy and value networks, followed by three independent layers of size 128, 128 and 64, where all the activation functions are taken to be ReLU.

\begin{figure}[t]
    \centering
    \includegraphics[width=\textwidth, trim={0 4cm 0 0},clip]{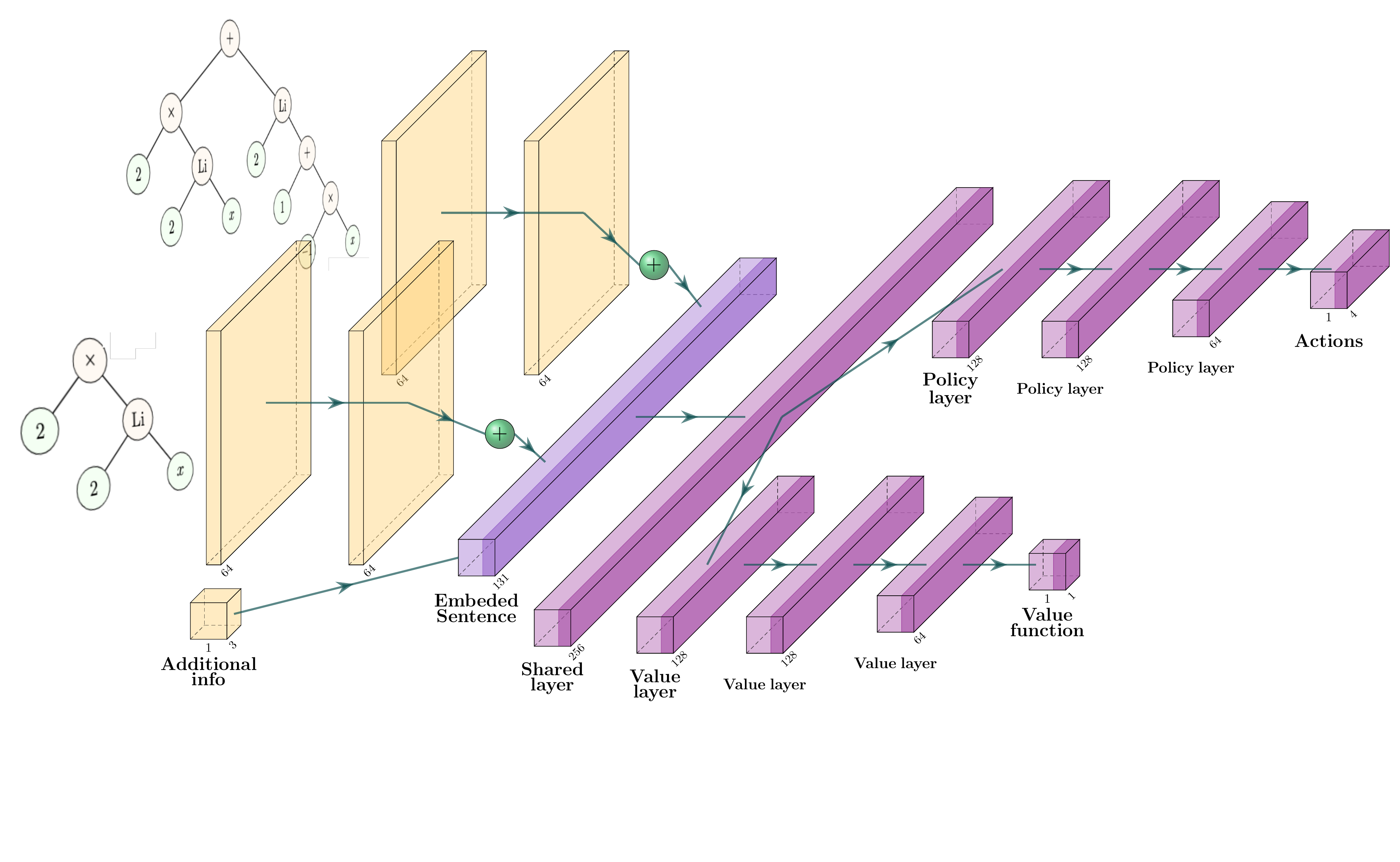}
    \caption{Architecture used for the embedding along with the policy and value networks.  We include the representation of the sentence embedding layer when a Graph Neural Network is used with two message passing layers. The full equation tree along with the first dilogarithm term are passed through distinct embedding channels. We apply a global pooling layer over all the nodes of the final layer to obtain a graph level embedding. The policy and value networks are composed of regular fully connected layers.}
    \label{fig:architecture}
\end{figure}

For the graph neural network used in the sentence embedding layer we refer to the GraphSAGE message passing architecture from \cite{graphsage} along with a mean aggregation scheme. We do not use a neighbor sampler over the nodes here contrary to GraphSage and have no LSTM aggregator. Our model is composed of 2 message passing layers and an embedding dimension of 64. Other architectures were considered such as Graph Convolution Networks from \cite{gcn} or Graph Attention Networks from \cite{gat}, but the simple linear message passing ended up giving the best performance. The complete architecture, including a GNN for the sentence embedding layer,  is illustrated in the Fig.~\ref{fig:architecture} with two message passing layers.

\bigskip
\noindent
\textbf{Episode}\quad An episode is defined by letting the agent attempt to simplify a given expression. At each episode reset we  sample a new expression to simplify, drawing from a previously generated testing set. During an episode the agent  runs for $n_{\text{steps}}$ steps at most, where each step corresponds to taking one action. Since we may not know how simple the final equation can be, we terminate the episode early only if the mathematical expression is reduced to 0. We will use $n_{\text{steps}}=50$ for training, although we also experimented with lengthier episodes. For simple examples increasing $n_{\text{steps}}$ does not necessarily lead to a better performance as taking more actions in the wrong direction will typically increase the length of the expression drastically. 

\subsection{Training the RL agent}
Our implementation is done in \texttt{Python} using standard  libraries such as \texttt{SymPy} \cite{sympy}, \texttt{Gym} \cite{gym} and \texttt{PyTorch} \cite{pytorch}. If no graph neural networks are used the architecture is simple enough that the RL agents can be trained on a standard CPU. The environment corresponding to our problem is made to match the requirements of a \texttt{Gym} environment allowing us to use the \texttt{stable-baselines3} \cite{stable-baselines3} RL library.

\subsubsection{Solving for different starting points}\label{sub-sec:rldata}
The objective of our RL algorithm is to reduce various linear combinations of dilogarithms. To ensure a simpler monitoring of the training we will consider the simpler sub-problem of simplifying expressions that are reducible to 0 after a suitable combination of identities. We create a generator for such expressions, making sure that our RL agent encounters a variety of different environments during training.

 \medskip
 \noindent
 \textbf{Generating starting points}\quad We generate our multiple starting points using a random function generator and a ``scrambling" procedure. By a scramble we refer here to the action of applying an identity to a given dilogarithm term in order to make it appear more complicated. The steps for creating a training expression are as follows
 \begin{enumerate}
        \item We sample the number  $n_t \in [1, 4]$ of distinct dilogarithms terms that we want to consider.
     \item  For each term we sample a single-variable rational function $h_i(x)$. We limit the degree of each $h_i(x)$ to be at most two, since the complexity of the expressions can increase drastically with the usage of the duplication identity. The coefficients of $h_i(x)$ are also constrained between $[-2,2]$, leading to around 7,000 unique rational functions. 
     \item For each dilogarithm term we sample an integer constant $c$ between $[1,8]$. At this point we can create a combination of dilogarithms that is explicitly 0 as 
\begin{equation}
     0 = \sum_{i=1}^{n_t}\,  c_i \, \Li_2(h_i(x)) -  c_i \, \Li_2(h_i(x))
     \end{equation}
     \item We sample the total number of scrambles to be applied $n_{scr}$, with $n_{scr} \in [n_t, 7]$. 
     \item We choose how to distribute the scrambles/identities amongst each term. We ask that every zero added, indexed by $i$, is acted upon by at least one identity.
     \item We apply the identities on $c_i \Li_2(h_i(x))$, making sure that no identity is applied twice in a row on the same term (or we would get back to the starting point).
     \item We discard any expression, which, when converted to prefix notation, is larger than $L_{\text{max}}$. This ensures that every example can fit inside the network.\footnote{In practice we observe that above $n_t =4$ or $n_{scr}=7$ some expressions start to reach 500 words in length. For $n_{scr}\in[5,6,7]$ we limit $n_t \in[1,2]$ to prevent this and keep simpler expressions.} In our architecture we take $L_{\text{max}}=512$.
 \end{enumerate}
 
 \noindent
There are around $10^{19}$ possible distinct expressions that can be produced this way.

\medskip
\noindent
\textbf{Training set} \quad Equipped with our generator we can create sets of starting points for the environment. We tested two different approaches: at each reset (each new episode) we can either create a brand new example or draw one from a previously generated set. We observed that both procedures gave similar results and find it easier to present the second case only. We generate an initial training set of 13,500 expressions, where each equation is annotated with the number of scrambles used to generate it. At each episode reset a new expression will be sampled from that set and used as a new starting point.  To give a few concrete examples, some of the equations considered are:
    \begin{align*} 
(n_{scr}=2, n_t=2) :  &\quad -2\, \text{Li}_2\left(\frac{-x^2+x+2}{2}\right)-7 \text{Li}_2\left(-2 x^2\right)-7 \text{Li}_2\left(-\frac{1}{2 x^2}\right)-2 \text{Li}_2\left(\frac{x^2-x}{2}\right)
\\[0.5em] 
(n_{scr}=3, n_t=1) :  &\quad \frac{5}{2}\, \text{Li}_2\left(\frac{1}{x^2}\right)-5 \, \text{Li}_2(1-x)-5 \, \text{Li}_2\left(-\frac{1}{x}\right)
\\[0.5em]
(n_{scr}=4, n_t=3) :  &\quad -\frac{3}{2} \, \text{Li}_2\left(\frac{1}{4 x^2-4 x+1}\right)-7\,  \text{Li}_2\left(\frac{2 x+2}{x+2}\right)-7\,  \text{Li}_2\left(-\frac{x}{x+2}\right)\\[0.2em]&\quad  -7 \, \text{Li}_2(2 x-1)-3\,  \text{Li}_2(1-2 x)-4\,  \text{Li}_2(2-2 x)
\end{align*}

\noindent
which can all be simplified to 0 (up to logarithms). In addition to the training set, we also generate a test set of 1,500 expressions, independent of the training set, which the agent will not see at all during training. 

\subsubsection{Details of the training runs}

\textbf{Hyperparameter tuning} \quad  We performed a non-exhaustive search for the tuning of our RL agent hyperparameters, starting with the defaults of \cite{stable-baselines3}. During the tuning the agents were set to run for $5\times 10^5$ time steps and the hyperparameters were picked in order to both maximize the expected rewards and minimize the mean episode length. The parameters that we ended up using are shown in Table~\ref{tab:hyper}. A finer optimization of the various parameters could be implemented but is not the main focus of this work. For the Graph Neural Network we used the default parameters of \cite{torch-geometric}.

\begin{table}[t]
\centering
{\renewcommand{\arraystretch}{1.25}%
\begin{tabular}{|c| c c|} 
 \hline
   Hyperparameter & TRPO Agent & PPO Agent  \\ 
 \hline
 
$n_{\text{steps}}$& 2048/4096\tablefootnote{We use 2048 steps when the sentence embedding layer is a GNN and 4096 otherwise.} & 4096  \\
Batch size& 128 & 128  \\
Learning rate& 10$^{-3}$ & 10$^{-4}$ \\
GAE $\lambda$& 0.9 & 0.9  \\
Discount factor $\gamma$& 0.9 & 0.9  \\
 \hline
Target KL $\gamma$& 0.01 & N/A  \\
Max steps in CG algorithm & 15 & N/A  \\
Number of critic updates & 10 & N/A  \\
 \hline
\end{tabular}}
 \caption{Hyperparameters used for the RL agents. For Generalized Advantage Estimation (GAE) we refer to \cite{gae}. The Kullback–Leibler divergence \cite{kl} constraint and the Conjugate Gradient algorithm \cite{Hestenes1952MethodsOC} are both used within TRPO \cite{trpo}.}
 \label{tab:hyper}
 \end{table}

\bigskip
\noindent
\textbf{Experiments} \quad We run two sets of experiments,
for the two choices of sentence embeddingg, Graph Neural Network or one-hot encoding, as described in Section~\ref{sec:rl-desc}. In both cases we will compare the performance of the agent with and without the cyclic penalty $\Delta r_t$  of Eq.~\eqref{eq:rew2} added to the reward function, for which we take $\lambda_r = 0.25$. For all experiments we let the agents run for 3M steps each, keeping a maximum episode length of $n_\text{steps}=50$.

\bigskip
\noindent
\textbf{Monitoring the training} \quad To get a simple measure of whether the agent is training correctly we focus on two different metrics: the average rewards and the mean episode lengths.
Since our equations can be simplified down to 0, at which point an episode is terminated early, we do expect to see the mean episode length decreasing with the training steps. To present readable plots we use a moving average with a window size of 20 policy iterations to smooth the learning curves. 

\begin{figure}[t]
    \centering
    \nocaption
    \begin{subfigure}[ht!]{0.48\textwidth}
\includegraphics[scale=0.5]{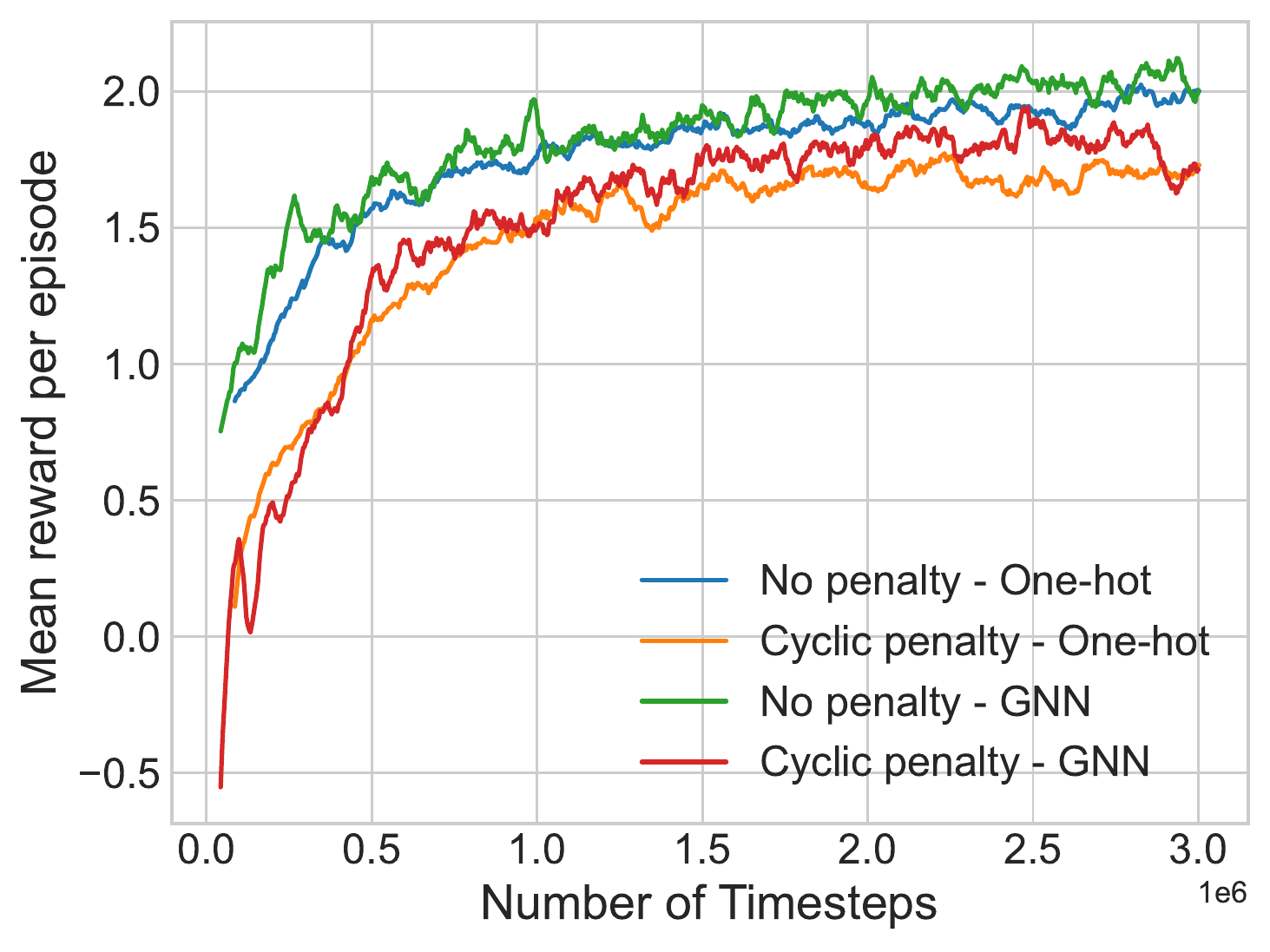}
    \end{subfigure}
    ~ 
    \begin{subfigure}[ht!]{0.48\textwidth}
      \includegraphics[scale=0.5]{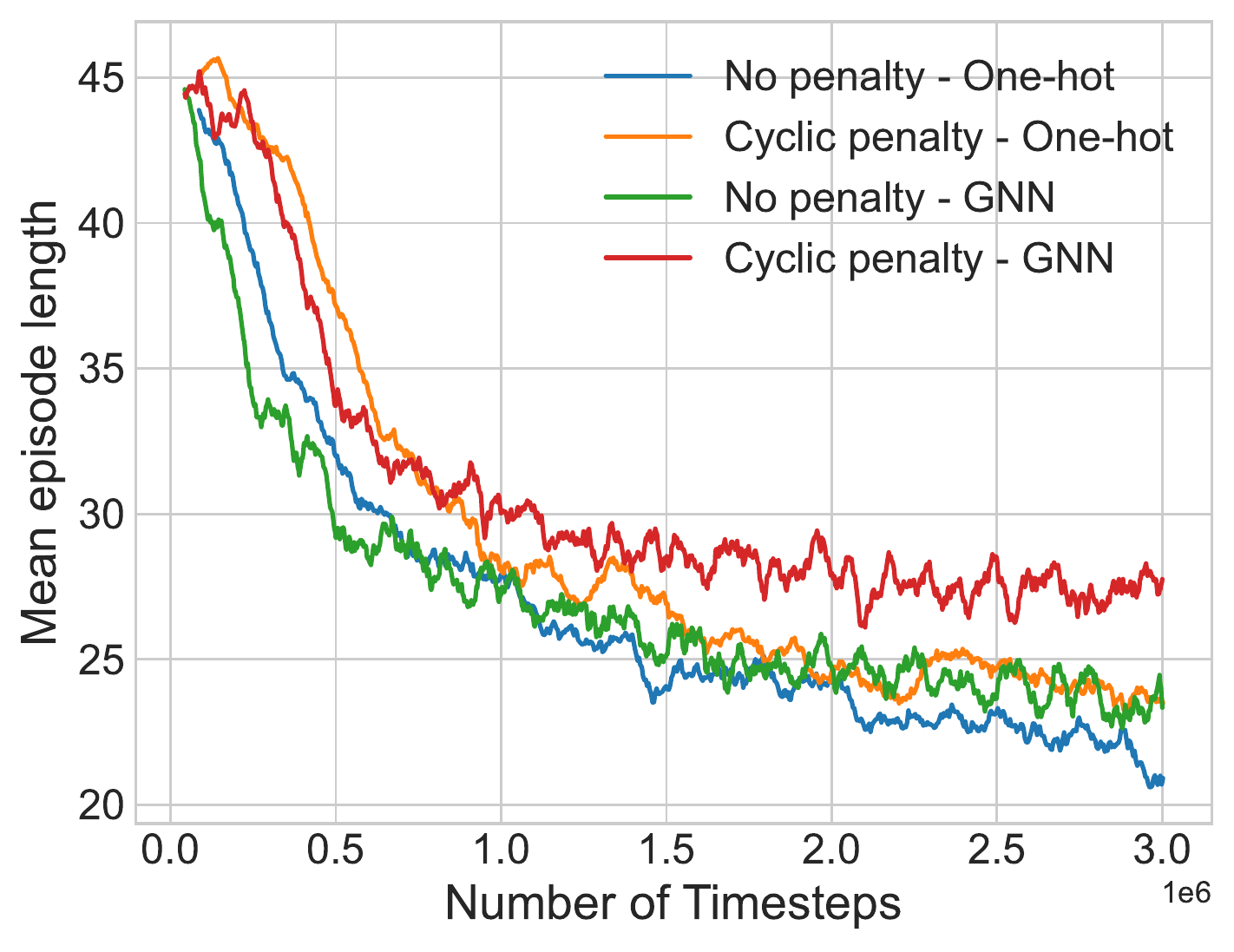}
    \end{subfigure}
    \caption{Learning curves of the various RL agents. Different embedding schemes are considered with either  one-hot encoding or a Graph Neural Network. The reward function without penalty is in Eq.~\eqref{eq:rew1}, while the one with a cyclic penalty refers to Eq.~\eqref{eq:rew2}. \label{fig:learning} 
    }
\end{figure}

The results of the training runs are shown in Fig.~\ref{fig:learning}. We observe that both sentence embedding schemes follow a similar learning curve; in both cases the episode length does decrease and reach comparable values. We also observe that the training involving the GNN is less stable, with the mean rewards fluctuating a lot more than in the one-hot encoding scheme.  As expected, the addition of the cyclic penalty to the reward function is reflected in Fig.~\ref{fig:learning} by a smaller total reward, especially for the early episodes. The fact that the cyclic penalty reward curves do not quite catch up to the base reward curves is an indication that our final agent can still get stuck in cyclic trajectories. Additional reward shaping was considered to minimize this effect but did not end up being beneficial for the overall performance.

\bigskip
\noindent
\textbf{Evaluating performance} \quad To accurately reflect how well our agents learned to simplify the starting expressions we perform a dedicated test run. For this we use our independent test set
 of 1,500 expressions. To probe an agent's performance we measure the number of equations that it is able to resolve, along with the number of identities that are required. We do not count the cyclic permutation action as using an identity here, since it does not correspond to an actual mathematical operation on the dilogarithms. Our goal is to quantify how much of the ``simplification tree" we have to explore compared to classical algorithms. Since all of the examples in the test set are tagged with the number of scrambles used to generate them, we can study the performance as a function of the complexity of the starting point.
 
 In order to reduce the expressions with our trained RL agents we try out two different approaches. Our first approach is a greedy, single-roll out one, where at each time step $t$ we perform the top action proposed by the policy network, taking 
 \begin{equation}
     a_t = \underset{a_t}{\text{argmax}} \, \pi_\theta(a_t|s_t) \, .
 \end{equation}
 Our second approach is a beam search one, where at every step we consider the top two actions proposed by the policy network. In order to limit the number of trajectories to explore, we only ever keep the top $N$ trajectories in memory. To determine which trajectories are better, we retain the ones that are expected to give us to the best return. This is measured by looking at
 \begin{equation}
     \sum_{k=0}^{t-1} \gamma^k r_k +\gamma^t r_t + V(s_{t+1})\, , 
 \end{equation}
summing the rewards already accumulated, the reward associated with the action $a_t$ and the reward to go, which we estimate with the value network. Here, $\gamma$ is the discount factor $\gamma$ mentioned in Table \ref{tab:hyper}.
 
\textbf{RL results}\quad We report in the Table~\ref{tab:rlres} the overall performance of each RL agent on the test set. We include as a benchmark the performance of a random agent and of a classical algorithm. For the classical algorithm, we use the modified best-first search algorithm, described in Section \ref{sec:mbfs}, which is left to explore the simplification tree up to a depth of 10. There is no major difference to be observed between both of our embedding schemes, with the GNN slightly outperforming the simple one-hot encoding approach. Overall we do notice that the inclusion of a cyclic penalty does help the agent, prompting it to explore a more relevant part of the state space during the exploration phase.


\begin{table}[t]
\centering
{\renewcommand{\arraystretch}{1.25}%
\begin{tabular}{l l c c c c} 

\toprule 
    Reward & Agents & \multicolumn{2}{c}{Greedy} & \multicolumn{2}{c}{Beam size 3}\\

    && Solved (\%) &   \begin{tabular}{@{}c@{}}\# Unscramble \\[-0.5em] steps\end{tabular}   & Solved (\%) &\begin{tabular}{@{}c@{}}\# Unscramble \\[-0.5em] steps\end{tabular} \\  
    \midrule[1pt]
    \multirow{2}{*}{No penalty}&One-hot&50 \%&5.3&78 \%&14.7\\
    &GNN&56 \%&6.4&80 \%&15.9\\
         \midrule
            \multirow{2}{*}{$\Delta r_t$ penalty} &One-hot&59 \%&7.4&85 \%&19.2\\
     &GNN&53 \%&8.7&\textbf{89 \%}&\textbf{20.3}\\  
         \midrule[0.9pt]
         &Random &13 \%&8.7&&\\
         &Classical &\textbf{91} \%&\textbf{39.3}&&\\
    \bottomrule[1pt]
    
\end{tabular}}
\caption{Performance of the trained RL agents and relevant benchmarks at reducing linear combinations of dilogarithms using functional identities. The number of unscramble steps is defined as the number of nodes of the ``simplification tree" that were visited for the solved expressions. The classical algorithm is the modified best-first search of Section \ref{sec:mbfs} with a depth of 10.}
\label{tab:rlres}
\end{table}

Although the agents do manage to learn in this challenging environment, with the greedy single-roll out evaluation they are only able to resolve $50 \sim 60 \%$ of the test equations. On the other hand, the beam search analysis is able to give more accurate results (89\%), indicating that the policy network's top prediction is not confident. It is important to correlate the performance with the number of evaluations (unscrambling steps) explored when searching for the simplified form, giving a  measure of how efficient the agent or algorithm is. In that regard the classical algorithm requires an larger exploration of the simplification tree before reaching the simplified form. That is, the classical algorithm essentially searches blindly while the RL algorithm searches in a directed manner. 

\begin{figure}[t]
    \centering
    \includegraphics[width=1\textwidth]{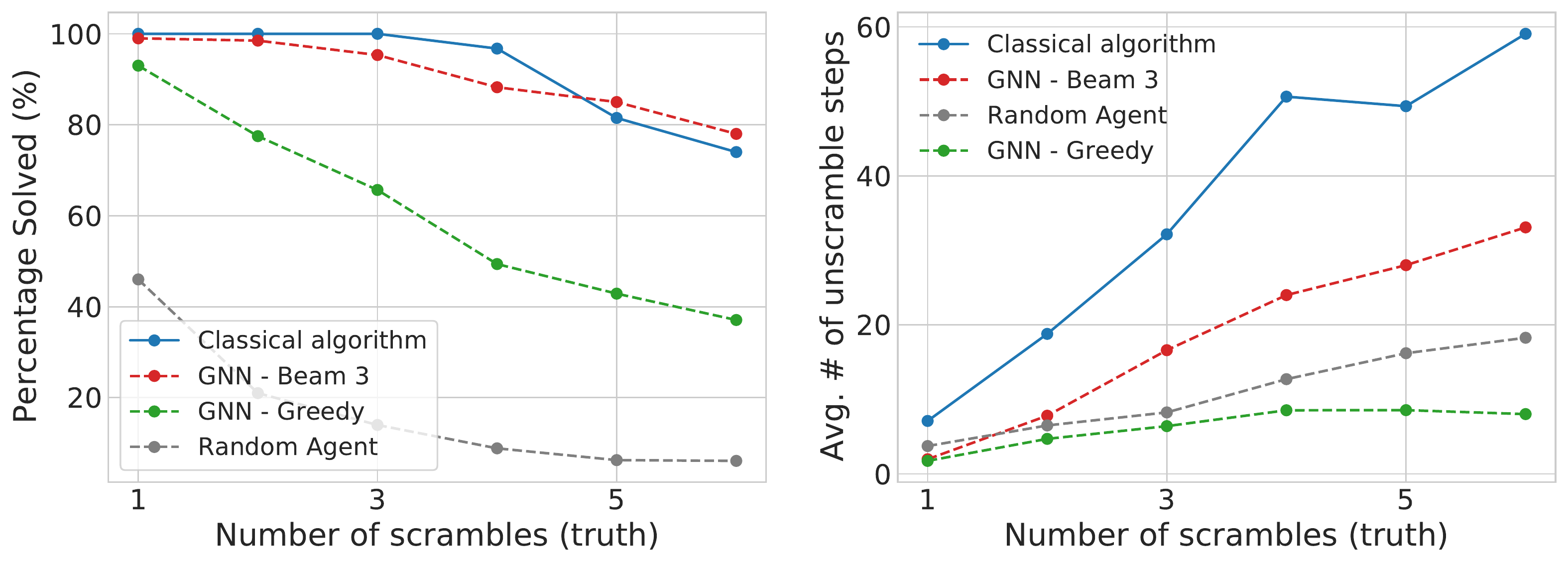}
    \caption{Performance of the best agents on the test set of 1,500 dilogarithmic expressions that simplify to 0, as a function of the complexity of the input, given by the number of identities (scrambles) applied to generate the expression. The right panel shows the average number of identities tried (unscrambling steps) by the different agents when solving the examples. The classical algorithm has worse scaling because it does essentially a blind search.
     }
    \label{fig:allsolved}
\end{figure}

We can also ask about the scaling of performance with the complexity of the input equation. To make this connection more precise we plot in Fig.~\ref{fig:allsolved} the performance as a function of the number of identities used (scrambling steps) to generate our expressions. Also shown is the number of identities used (unscrambling steps)  when looking for the simplified form. 
The greedy RL agents only consider a single action per step, which is inexpensive, and so the scaling observed is highly correlated with the minimal number of steps required to solve an expression. When using a beam search of size $N$ we can take up to $2 N$ evaluations per step, since we are probing the two best actions for the $N$ trajectories in memory. However, even for the beam search the number of unscrambling steps explored does not scale with the expression's length but only with the number of steps required to find the solution. This is to be contrasted with the classical algorithm, where the number of evaluations required per step scales with the number of actions and the number of distinct dilogarithm terms. For expressions that have been scrambled a lot, which tend to be lengthier due to the duplication identity,
the number of evaluations required by the classical algorithm quickly becomes unmanageable. We can see a realization of this behaviour on the Fig.~\ref{fig:allsolved} where the classical algorithm needs to perform a much broader search before finding the solution. The  performance for all of the agents can be found in Appendix~\ref{app:perf}.

\medskip
\noindent
\textbf{Challenges of RL} \quad  Our trained agents have learned  to simplify linear combinations of dilogarithms, reaching a beam search performance of $89 \%$ compared to the $13 \%$ of the random agent. However the environment considered was fairly straightforward since the expressions considered were the ones reducing to 0. Training in the general case, where the goal state is not known, is expected to be much harder. Nonetheless RL seems to offer some advantages over at a simple classical algorithm, for instance being able to reduce the number of evaluations required to find the simplified expression. 

One can look to analogous problems for possible improvements. For example, a classic RL problem is solving the Rubik's cube.
In~\cite{deepcube} classical algorithms were used in conjunction with policy and value networks to guide the search for the shortest path to the solution. 
Even though solving the Rubik's cube and simplifying dilogarithms share similarities as a search problem the nature of the two environments are somewhat different. In our symbolic setting the representation of the environment itself must use various embedding layers, the action space does not have all actions on the same footing (the duplication identity tends to increase the length of expressions if used incorrectly for instance), and the trajectories taken to reach the correct solution are generally heavily constrained (these trajectories are generically unique, up to a cyclic usage of actions). Training the embedding layers jointly with the policy and value networks also proved complicated, as is apparent by the fact that using a GNN did not yield significant gains over the naive encoding approach. As was done in \cite{unknot}, one could try out larger networks like the reformer \cite{reformer}, known to provide a robust semantic embedding. However we find that in practice it is hard to train these large architectures in conjunction with the policy and value networks. In fact it has been observed that in typical RL settings bigger networks are not necessarily correlated with better performance \cite{deeprl, largenet}. One strategy adopted by \cite{largenet} to tackle this issue is to decouple the representation learning from the actual RL by pre-training the embedding layer for the state. However such an approach is not easily applicable in our context. Indeed whereas powerful sentence embedding schemes exist for natural language \cite{sentembed} they do not seem adapted for representing mathematical data. One could also consider pre-training a graph embedding model \cite{graph2vec} on a subset of well chosen equation trees. The problem of simplifying polylogarithms with RL is a rich one and worthy of continued exploration.

\section{Transformer Networks \label{sec:transformer}}
In Section~\ref{sec:RL} we explored how reinforcement learning can be used to train agents to simplify linear combinations of dilogarithms by using various functional identities. An advantage of the RL approach is its reproducibility: the list of actions taken to arrive at the final answer is known and can be studied and checked.
However, demanding reproducibility also imposes a limitation on the way the data is learned and understood. RL algorithms such as ours can suffer from sample inefficiency as commonly seen in other applications of policy gradient methods~\cite{GuLilGhaTurLev17}: since they only train on the current version of the policy, they can be slow to train.  This slowless limited our training sample to only 13,500 expressions. It is actually somewhat remarkable that the RL network can perform so well with so little diversity in the training. Partly this is due to the RL algorithm devoting 50 time steps to each expression during training.

If we are only interested in the final simplified form, not the explicit set of identities used to get there,
 \texttt{seq2seq} models are well adapted, with the prime example being the Transformer Network~\cite{transformer}.
 A transformer network will generate a guess for the answer, which we can verify, but not a human-interpretable sequence of steps to get there. Our approach is based on the work of \cite{fbtransformer}, where transformer networks were used to perform a symbolic integration task, showing competitive performance compared to classical algorithms used by \texttt{SymPy} or $\texttt{Mathematica}$. 

To use the transformer we have to convert our task of simplifying equations into a translation one. We will explore two different approaches. In Section \ref{sec:transfo-dilogs} we will inquire whether it is possible to train a transformer to  simplify linear combinations of dilogarithms directly, as in the problem RL was applied to in Section \ref{sec:RL}. Then in Section \ref{sec:transfo-symbol} we will explore using the symbol to represent a polylogarithmic expression and then applying the transformer network to look for a simplest expression that is associated with a given symbol.

From a high level point of view the transformer takes as input a sentence, in our case a mathematical expression, encodes it, and proceeds to output the translated sentence. As with the RL approach described in Section~\ref{sec:RL}, the input data in our case are mathematical expressions which we express as trees and parse into prefix notation. The particularity of the transformer lies in that both the encoding and the decoding process rely on the attention mechanism, ensuring that any given word knows about the other constituents of the sentence. In the decoding layer the attention mechanism is used both on the input sentence and on the words of the translated sentence that have already been generated. We refer to \cite{transformer} for a detailed overview of the architecture.

The particular attention mechanism we use is called ``Scaled Dot-Product Attention". We start by passing each word through an embedding layer, representing it as a vector in $\mathbb{R}^{d_{\text{emb}}}$, with $d_{\text{emb}}$ the embedding dimension. For a given word w$_i$ the resulting embedding $w_i$ is associated with a query vector $q_i$, a key vector $k_i$ and a value vector $v_i$. Those are obtained by multiplying the word embedding with suitable weight matrices, learned during the training. For a given attention head we have $q_i = W^Q w_i$, where $W^Q \in \mathbb{R}^{d_{\text{emb}} \times d_h}$, with $d_h$ the dimension of the attention head. When using multi-headed attention with $h$ heads we will take $d_h= d_{\text{emb}}/h$, following the implementation of \cite{fbtransformer, transformer}. To calculate the attention for a given word $w_i$ we compute the dot product $q_i \cdot q_j$ with $j$ going over all the words in the sentence. In practice this quantifies how similar $w_i$ is to the other words of the sentence. After dividing by $\sqrt{d_h}$ we apply the softmax function to obtain the attention weight $\alpha_{i, j}$  for the word $w_i$ associated with the word $w_j$:
\begin{equation}
    \alpha_{i, j} = \frac{\exp(a_{i, j})}{\sum_k \exp(a_{i, k})} \quad \text{with} \quad a_{i, j} = \frac{q_i \cdot q_j}{\sqrt{d_h}}
\end{equation}

\noindent
The final output $o_i$ of the attention layer for the word $w_i$ is given by multiplying the attention weights with the value vectors 

\begin{equation}
o_i = \sum_j \alpha_{i, j}\,  v_j \,
\end{equation}
and the new embedding learned is dependant on all of the other words in the sentence. To remain sensitive to the ordering of the words we have to add a positional encoding layer at the start of the model, following \cite{positional_embed}.

\subsection{Simplifying dilogarithms with the transformer} \label{sec:transfo-dilogs}


Having introduced the transformer network, we first deploy it to tackle a problem similar to the one we have motivated in Section \ref{sec:RL}.

\bigskip
\noindent
\textbf{Data generation} \quad Since we are using the transformer to guess the answer, we will have to consider dilogarithmic expressions which do not necessarily simplify to 0 (otherwise the prediction will be trivial and no learning will take place). In order to create relevant training samples we slightly modify the generation script of Section \ref{sub-sec:rldata}. To create a scrambled dilogarithmic expression we proceed in the following way
\begin{enumerate}
    \item We sample the number of distinct dilogarithms $n_s \in [0,3]$  for the simplified expression.
    \item We sample the number of times $n_t \in [0,3]$ that we want to add zero in the form $c_i\, \Li_2(h_i(x)) -c_i\, \Li_2(h_i(x))$.
    \item We create random arguments $g_i(x)$ and $h_j(x)$ for each dilogarithm. Each function is a rational function of degree 2 at most over the integers between $-2$ and $2$.
  This gives us  a skeleton 
    \begin{equation}\label{eq:scr}
    f(x)=\sum_{i=1}^{n_s}\, a_i \, \Li_2(g_i(x)) + \sum_{j=1}^{n_t}\left[c_j\, \Li_2(h_j(x)) -c_j\, \Li_2(h_j(x)) \right] 
\end{equation}

    \item We sample the total number of scrambles (inversion, reflection or duplication) to do. Here we take up to 10 scrambles as the scrambles may be applied on either the $g_i$ or the $h_j$ terms of Eq.~\eqref{eq:scr}.
    We are able to allow for more scrambling with the transformer network than the RL network in Section \ref{sub-sec:rldata} (which allowed for only 6-7 scrambles) because the transformer network is easier to train.
    \item We randomly choose how to distribute the scrambles amongst each term. We ask that every zero added, indexed by $j$, is scrambled at least once.
    \item We apply the scrambles and discard the resulting expression if it has more than $512$ words. 
\end{enumerate}

\smallskip
Training the transformer is done in a supervised way: for each input expression $f_i$ we have an associated output expression $F_i$ which we try to recover. In our case the input expressions are the scrambled expressions, while the desired outputs will be the corresponding simple expressions of the form $\sum_i\, a_i \, \Li_2(g_i(x))$. Following the outlined data generation procedure we create about 2.5M distinct pairs $\{f_i, F_i\}$ to be used for the training and retain 5,000 examples for testing the performance.\footnote{Note that we can have a training set for the transformer roughly 100 times larger than for the RL network since the RL network visits each example multiple times.}
We also make sure that no scrambled equation from the test set can be found in the training set. Table~{\ref{tab:eg1}}  gives some examples from the test set. It is important to remember that we are only trying to show the equivalence between two expressions up to $\ln^2, \ln$ and constant terms.

 \begin{table}[htp!]
       \centering
       {\renewcommand{\arraystretch}{3.0}%
\begin{tabular}{|c|c|c|}
\hline
        Scrambled expression $f_i$ & Simple expression $F_i$   \\
        \hline 
        \hline 
        $-4\, \text{Li}_2\left(\frac{4}{x^2-8 x+16}\right)+8\, \text{Li}_2\left(\frac{2}{x-4}\right)-8\, \text{Li}_2\left(\frac{2}{x-2}\right)$ & $0$\\[0.3em]
               \hline 
        $\begin{array}{l}
       -6 \, \text{Li}_2\left(\frac{1}{x^2+2 x}\right)-6\, \text{Li}_2\left(x^2+2 x\right)+\frac{7}{2}\, \text{Li}_2\left(\frac{4}{x^2-4 x+4}\right)\\[-1em]
     ~~~~~ -7\, \text{Li}_2\left(\frac{x}{2}\right)-7\, \text{Li}_2\left(\frac{2}{x-2}\right) \end{array}$& $0$\\[0.3em]
               \hline 
$\begin{array}{l}
      -8 \,\text{Li}_2\left(\frac{x^2}{2 x-2}\right)-8\, \text{Li}_2\left(-\frac{x^2}{2 x-2}\right)-\frac{7}{2}\, \text{Li}_2\left(\frac{4}{x^2+2 x+1}\right)\\[-1em]
      ~~ +4\, \text{Li}_2\left(x^2+4 x+4\right) +4\, \text{Li}_2\left(\frac{x^4}{4 x^2-8 x+4}\right)-7\, \text{Li}_2\left(-\frac{x}{2}-\frac{1}{2}\right)\\[-1em]
      ~~~~~-2\, \text{Li}_2\left(\frac{1}{x+2}\right)+8\, \text{Li}_2\left(-\frac{1}{x+2}\right)-10\, \text{Li}_2(x+2)\end{array}$& $-7 \, \text{Li}_2\left(\frac{2}{x+1}\right)$\\[0.3em]
               \hline 
               $\begin{array}{l}  -4 \, \text{Li}_2\left(x^2-x+1\right)+3\, \text{Li}_2\left(x^2-2 x+1\right)-3 \, \text{Li}_2\left(\frac{1}{x^2-4 x+4}\right)\\[-1em]
               ~~+\frac{5}{2}\, \text{Li}_2\left(\frac{x^2+2 x+1}{4 x^2+4 x+1}\right)-4 \, \text{Li}_2(-x-1)-5\, \text{Li}_2\left(\frac{x+1}{2 x+1}\right)\\[-1em]
               ~~~~+2 \,\text{Li}_2\left(x^4-2 x^3+3 x^2-2 x+1\right)+6 \, \text{Li}_2\left(\frac{1}{x-2}\right)-4\, \text{Li}_2(x+2)\\[-1em]
               ~~~~~~-3\, \text{Li}_2\left(\frac{1}{2 x+1}\right)-5 \, \text{Li}_2\left(\frac{-x-1}{2 x+1}\right)-3 \, \text{Li}_2\left(\frac{2 x}{2 x+1}\right)\end{array}$&
                $\begin{array}{l}
               6\, \text{Li}_2(1-x)\\[-1em]
               ~~-4\,  \text{Li}_2\left(-\frac{1}{x^2-x+1}\right)
               \end{array}$
               \\[0.3em]
              \hline 
               $\begin{array}{l}
      -6 \, \text{Li}_2(\frac{1}{2}\, -x)+8\, \text{Li}_2\left(-\frac{x^2}{2}-\frac{x}{2}\right)+\text{Li}_2\left(x-x^2\right)\\[-1em]
      ~~-8\, \text{Li}_2\left(\frac{1}{x^2-2}\right)-3\, \text{Li}_2\left(x^2+\frac{x}{2}+1\right)+3\, \text{Li}_2\left(\frac{2 x^2+x}{2 x^2+x+2}\right)\\[-1em]
      ~~~~-\frac{1}{2}\, \text{Li}_2\left(-x^4+2 x^3-3 x^2+2 x\right)-8\, \text{Li}_2\left(\frac{1}{2 x-2}\right)\\[-1em]
      ~~~~~~-8\, \text{Li}_2(2 x-2)-6\, \text{Li}_2\left(-\frac{2}{2 x-1}\right)\end{array}$& $\begin{array}{l}-8\, \text{Li}_2\left(\frac{x^2}{2}+\frac{x}{2}+1\right)\\[-1em]
      ~~+\text{Li}_2\left(-x^2+x-1\right)\\[-1em]
      ~~~~+8\, \text{Li}_2\left(x^2-2\right)\end{array}$\\[0.3em]
               \hline 
   \end{tabular}}
   \caption{Examples of data pairs $\{f_i, F_i\}$ found in the test set for the dilogarithm simplification task. Equations are generated following the procedure detailed in section \ref{sec:transfo-dilogs}. Arguments are simplified using the \texttt{cancel} function of \texttt{SymPy}. All examples are solved by the transformer.}
   \label{tab:eg1}
   \end{table}

\bigskip
\noindent
\textbf{Training parameters} \quad For our transformer model architecture we used 3 encoder and decoder layers, where the multi-headed attention blocks were composed of 4 heads. The embedding dimension was taken to be 512 and the positional encoding were obtained via a dedicated embedding layer (no sinusoidal embedding as in \cite{transformer}). Using a batch size of 32 equation pairs we trained for 200 epochs, with 50,000 expressions per epoch. This corresponded to a training time of 8 hours on a single A100 GPU. In line with the comments of \cite{fbtransformer} we observed that the overall performance of the model was not highly sensitive to the choice of hyperparameters of the transformer.

\bigskip
\noindent
\textbf{Assessing performance} \quad To determine how well the transformer is performing it is not sufficient to  consider only the accuracy observed during training. This accuracy metric is computed by looking at the words predicted by the transformer. However two mathematical expressions could be equivalent but have a different prefix notation. For that reason it is necessary to perform a dedicated evaluation run, where performance is assessed by looking at the actual mathematical expression predicted.

 \smallskip
From the point of view of our data generation procedure, expressions linked by reflection and inversion share the same complexity and can lead to different correct answers for the same prompt. For instance, looking at the second example of Table~\ref{tab:eg1} we would say that both $-7 \, \text{Li}_2\left(\frac{2}{x+1}\right)$ and $7 \, \text{Li}_2\left(\frac{x+1}{2}\right)$ are valid answers. To check such an equivalence we use the symbol. Since our model is blind to $\ln^2$ terms, two expressions are considered equivalent provided they share the same symbol, up to symmetric parts. Indeed any symmetric symbol will correspond to a $\ln^2$ term:
\begin{equation}
    \cS\left[\ln(g_1(x)) \ln(g_2(x))\right] = g_1(x) \otimes g_2(x) +  g_2(x) \otimes g_1(x) \,.
\end{equation}

\noindent
Thus when comparing the true answer $F_1(x)$ to the transformer's guess $F_2(x)$ it will be sufficient to assess
\begin{equation}
   A(\cS[F_1(x)] ) - A(\cS[F_2(x)] ) \stackrel{?}{=} 0
\end{equation}
where $A$ refers to taking the antisymmetric part of the symbol
\begin{equation}
    A(g_1(x) \otimes g_2(x)) = g_1(x) \otimes g_2(x) - g_2(x) \otimes g_1(x) \,.
\end{equation}

\noindent
In practice we check this equivalence by computing the symbol of the transformer's guess with the \texttt{PolyLogTools} package \cite{polylogtools}.

\subsubsection{Results}

\begin{figure}[t!]
    \centering
    \includegraphics[width=0.9\textwidth]{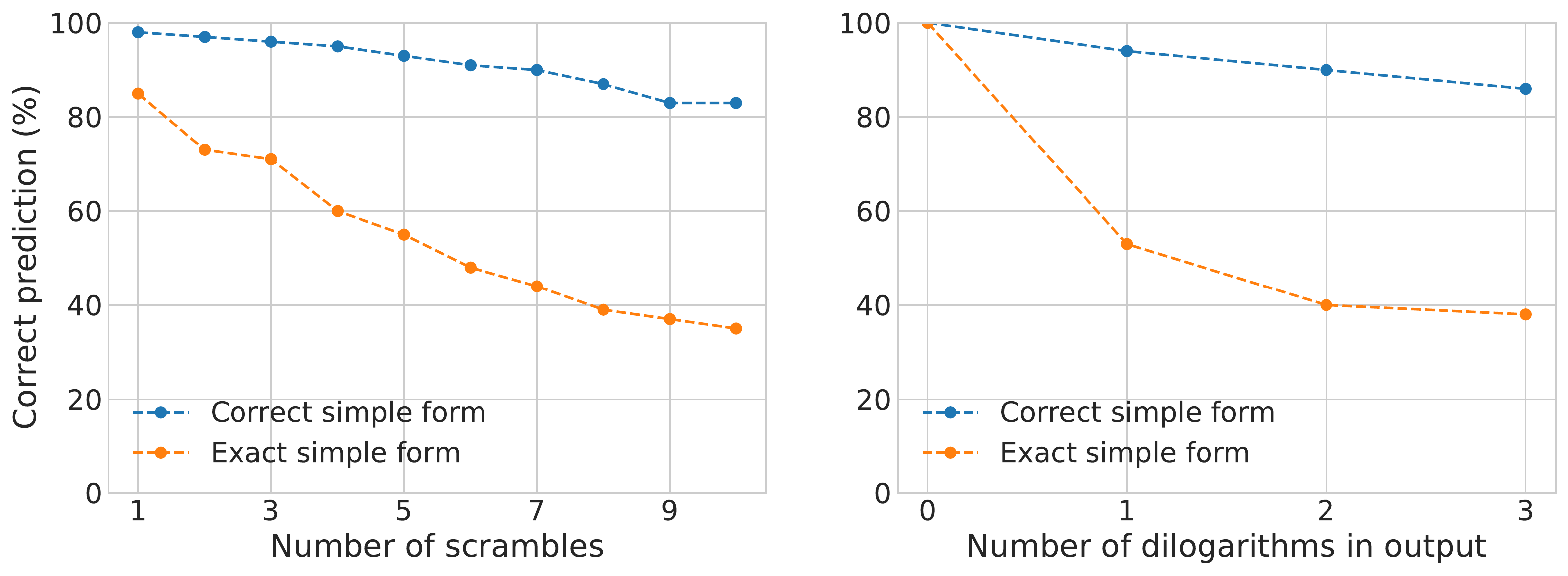}
    \caption{Performance of the transformer in simplifying expressions of dilogarithms as a function of two different measures of complexity. The left panel shows how performance depends on the number of identities (scrambles) applied to generate the expression. The right panel shows performance as a function of the number of terms in the simplified expressions. The networks predicts a unique answer which may be exactly the same as the simplified answer in the test set (orange points) or equivalent up to application of inversion and reflection identities which do not increase the complexity, and therefore still correct (blue points).}
    \label{fig:perf1}
\end{figure}

Using our trained transformer model we iterate through the test set and predict the simple expression corresponding to a given scrambled input. Since there are many equally simple forms, related by inversion and/or reflection, we should not expect the transformer to necessarily come up with the particular simple form in the test set. Indeed, simply comparing in \texttt{SymPy} the transformer's guess and the expected answer, gives a 53\%  accuracy. However, we can also check whether the prediction and the test set answer are related by inversion and reflection by comparing the antisymmetric part of the symbols. Doing so we find a 91\% accuracy. Thus the network is able to predict one of the "synonyms" of the correct answer quite effectively. Performance of the direct comparison and the equivalent comparison are shown in Fig.~\ref{fig:perf1} as functions of complexity. 

One concern is that because the performance evaluation on the test set is not exactly aligned with the loss used in training, the network might not be training as efficiently as possible. Unfortunately, our performance evaluation requires computing the anti-symmetric part of the symbol, which is too slow to be done in training. An alternative is to pick an element of the equivalence class of simple expressions by putting them in a canonical form. This can be done on both the training and test set.
To do so, we took our same training and test data sets and for each simple expression computed the six equivalent dilogarithm forms using inversion and reflection and then picked the simplest one according to some essentially arbitrary metric. Training and testing with this dataset the transformer had an accuracy of 92\%, comparable to the performance without choosing a canonical form. Thus, having multiple possible correct forms did not seem to hinder the network's training or performance.

   
   It is worth pointing out that the transformer does an excellent job in predicting the number of terms  that the expected answer should have. In fact, 97\% of the predictions contain the same number of distinct dilogarithm or logarithm functions as in the desired output.\footnote{In fact in some rare instances the transformer accurately predicts a number of terms lower than what is given as the ``true" output. This could happen when an expression is generated and could be simplified, but not in an obvious way. For instance we could generate $a \Li_2(x) + b \Li_2(1-x)$ where the simplest form would actually be $(a -b)\Li_2(x)$.}. In general, predicting the full answer gets harder with the length of the expected sequence, as shown in Fig.~\ref{fig:perf1}. We also notice that the performance is impacted by the number of scrambles applied to the input expression. This drop in performance can be associated with the complexity of the input since typically the number of scrambles applied is correlated with lengthier expressions, due to the presence of the duplication identity. 

The sharp difference between both our evaluation metrics can be understood better once we let the model predict a beam of answers as opposed to a single guess. For the beam search details we refer to \cite{fbtransformer, beam_search}. It is instructive to look at a particular example, for a beam of size 5, displayed in the Table~\ref{tab:beam5}. Looking at the predicted arguments, we can see that the transformer model is able to learn the different equivalent representations of a given function, where we can check that reflection and inversion identities link them all together. This redundancy has been effectively incorporated in the model, without any specific guidance from our part.  

 \begin{table}[htp!]
       \centering
       \begin{tabular}{ll}$
       \begin{array}{|ll|}
        \cline{1-2}
         & \\[0.2em]
            Input: & -\text{Li}_2\left(-2 x^2\right)+\text{Li}_2\left(-\dfrac{1}{2 x^2}\right)\\[1em] &
            ~~-7 \text{Li}_2(-2 x)-7 \text{Li}_2\left(-\dfrac{1}{2 x}\right)   
           \\[1em] & 
           ~~~~~4 \text{Li}_2\left(x^2-2 x+2\right)\\[1em]
           \cline{1-2} 
          & \\[0.5em]
            \textit{Target}:& 4 \text{Li}_2\left(-x^2+2 x-1\right) \\[1.5em]
            \cline{1-2} 
       \end{array}$&  $\Rightarrow \quad$ 
       {\renewcommand{\arraystretch}{3.0}%
\begin{tabular}{|c|c|c|}
\hline
        Hypothesis  & Valid ?  \\
        \hline 
        \hline 
       $-4 \text{Li}_2\left(x^2-2 x+2\right) $ & $ \checkmark^\star $\\
               \hline 
       $-4 \text{Li}_2\left(-\dfrac{1}{x^2-2 x+1}\right)$ & $ \checkmark^\star $\\
      \hline
       $4 \text{Li}_2\left(-x^2+2 x-1\right) $ & $ \checkmark $\\
      \hline
       $4 \text{Li}_2\left(\dfrac{1}{x^2-2 x+2}\right) $ & $ \checkmark^\star $\\
      \hline
       $-4 \text{Li}_2\left(-\dfrac{1}{x^2+2 x+1}\right)$ & $ \times $\\
      \hline
   \end{tabular}}
   \end{tabular}
   
   \caption{First 5 best hypothesis of the transformer for a given input. Hypothesis annotated with $\checkmark^\star$ are correct for a symbol based evaluation performed using \texttt{PolyLogTools}.}
   \label{tab:beam5}
   \end{table}

\subsection{Integrating the symbol with the transformer} \label{sec:transfo-symbol}

To evaluate the performance of the transformer model it was helpful to use the symbol. As discussed in Section \ref{sec:symbol} the symbol is a convenient tool for deriving identities and simplifying expressions. 
To simplify an expression with polylogarithms we can compute its symbol, simplify it using the product rule, and then integrate it back to give a polylogarithm. The simplification at the symbol level is typically easier to do since it follows the same type of identities as the logarithm. The problem is then reduced to one of integrating the symbol. No comprehensive classical algorithm is known for this task and we refer to Section \ref{sec:symbol-integration} for our version, designed to integrate symbols with two entries. For symbols with three entries or more (corresponding to polylogarithms with weight 3 or more) this is a much harder task and writing a classical algorithm is not straightforward. Working with the symbol allows us to go beyond the leading polylogarithmic part of the answer as well, so we will consider functions like $\ln^2(x)$ in addition to $\Li_2(x)$.

\bigskip
\noindent
\textbf{Training data} \quad Following the general discussion of Section \ref{sec:transfo-dilogs} we want to create a training and test set with pairs of symbols and functions $\{S_i, F_i\}$. Since computing symbols is straightforward in \texttt{Mathematica} with \texttt{PolyLogTools} we will first generate a set of appropriate functions $\{F_i\}$ and then compute their symbols to complete the dataset. We will restrict ourselves to functions $F_i(x)$ which are composed of classical polylogarithms with single-variable rational function arguments. When generating these functions we will assume that they have a uniform transcendental weight. The basis terms considered are listed in Table~\ref{tab:basis}.

 \begin{table}[t]
 \centering
\begin{tabular*} {.7\linewidth}{@{\extracolsep{\fill}}|ccc|}
 \cline{1-3}
   Weight 2 & Weight 3 & Weight 4  \\ [0.2em]
 \hline
  $\Li_2(x)$ &  $\Li_3(x)$&$\Li_4(x)$  \\ [0.1em]
  $\ln(x) \ln(y)$ &  $\Li_2(x) \ln(y)$ &  $\Li_3(x) \ln(y)$  \\ [0.1em]
   & $\ln(x)\ln(y)\ln(z)$&  $\Li_2(x) \Li_2(y)$  \\ 
   & &  $\Li_2(x) \ln(y) \ln(z) $  \\ 
   & &  $\ln(w) \ln(x) \ln(y) \ln(z) $  \\ 
 \hline
\end{tabular*}
\caption{Set of basis terms at a given transcendental weight.}
\label{tab:basis}
\end{table}

\noindent
Generating functions $F_i(x)$ is achieved by the following procedure:
\begin{enumerate}
    \item We sample a number of distinct terms $n_s \in[1,3]$. 
    \item For each distinct term we sample the overall coefficient $c_j$ and the type of basis term $B_j$ it corresponds to. This gives us a skeleton
    \begin{equation}
        F_i(x) = \sum_{j=1}^{n_s} c_j \, B_j(x)
    \end{equation}
    \item For each $B_j(x)$ we construct its argument(s) as rational functions of $x$ following the tree like generation of \cite{fbtransformer}. The nodes of the tree are either unary or binary operators, while the leafs are either constants or $x$. Numbers are represented using the numerical decomposition scheme of Section \ref{sec:rl-desc} with leaf constants allowed to take values from $-5$ to $5$. 
    
    \item If a logarithm is generated as part of a basis function then we preprocess it. We remove any overall constant and fully expand it.
    \item We discard any equations that have more than 512 words.
\end{enumerate}

\noindent
The complete set of parameters and ranges used can be found in Appendix~\ref{secap:data}. 

From each function $F_i(x)$ we then compute its symbol.
To increase the dataset diversity we can also apply some scrambling moves at the level of the symbol. These moves follow from the general product identity of the logarithm and are
\begin{align*}
     (\textit{Product rule}) :&\quad \cdots \otimes f(x) \otimes\cdots \quad \longleftrightarrow \quad  \cdots \otimes f(x) g(x) \otimes \cdots - \cdots \otimes  g(x)\otimes \cdots \\*[0.4em]
    (\textit{Quotient rule}) :&\quad  \cdots \otimes f(x)\otimes \cdots \quad  \longleftrightarrow   \quad \cdots \otimes \frac{f(x)}{ g(x)}\otimes \cdots + \cdots \otimes  g(x)\otimes \cdots \\*[0.3em]
    (\textit{Power rule}) : & \quad \cdots \otimes f(x)\otimes \cdots\quad  \longleftrightarrow  \quad  \cdots \otimes \frac{1}{d}\, \left(f(x)\right)^d \otimes\cdots
\end{align*}
where we impose $d \leq 3$ and take the functions $g(x)$ to be  sampled as polynomials of degree 3 at most. In the following we will limit the number of possible symbol scrambles to 5 at most.
In our scrambling procedure we also allow the reverse moves, if they exist. So for instance we allow for the expansion of $\cdots \otimes f(x) g(x) \otimes \cdots \rightarrow \cdots \otimes f(x)  \otimes \cdots + \cdots \otimes g(x)  \otimes \cdots$ if we have such an explicit factorization.
Some examples of pairs functions and symbols generated in this way are shown in Table~\ref{tab:eg2}. Symbols parsed with more than 1024 words are discarded.

 \begin{table}[htp!]
       \centering
       {\renewcommand{\arraystretch}{3.0}%
\begin{tabular}{|c|c|c|}
\hline
        Input symbol $\cS_i$ & Simple expression $F_i$   \\
        \hline 
        \hline 
        $\begin{array}{l}
      -\left(-x^2-x+1\right)\otimes (1-x)+\left(-x^2-x+1\right)\otimes  x \\[-1em] -\left(-x^2-x+1\right)\otimes (x+1)+x\otimes (1-x)-x\otimes x+x\otimes (x+1)\end{array}$& $\text{Li}_2\left(\frac{(1-x) (x+1)}{x}\right)$\\[0.3em]
               \hline 
$-\frac{23}{4}  (1-x)\otimes x-6 \left(x^2+x+1\right)\otimes x$& $2 \,\text{Li}_2\left(x^3\right)-\frac{1}{4}\, \text{Li}_2(x)$\\[0.3em]
               \hline 
               $\begin{array}{l}
     -40 \left(6-x^2\right)\otimes \left(6-x^2\right)-3 (1-x)\otimes \left(-x^6-x^2+3\right)\\[-1em]-3 (x+1)\otimes \left(-x^6-x^2+3\right)-3 \left(x^4+x^2+2\right)\otimes \left(-x^6-x^2+3\right)\\[-1em]+\frac{1}{4} (5-2 x)\otimes (2-x)\end{array}$& $\begin{array}{l}3 \text{Li}_2\left(-x^6-x^2+3\right)\\[-1em]-\frac{1}{4}\, \text{Li}_2(2 x-4)\\[-1em]-20 \ln ^2\left(x^2-6\right)\end{array}$\\[0.3em]
               \hline 
          $\begin{array}{l}
8 \, \dfrac{x^2-x-1}{x-1}\otimes x-8 \left((x+1) \left(x^2-x-1\right)\right)\otimes x\\[-1em]+8 (1-x)\otimes \left(-x^3+x^2-x-1\right)-8 \dfrac{1}{x-1}\otimes x \\[-1em]-8 (1-x)\otimes \left(x \left(x^3-x^2+x+1\right)\right)\end{array}$& $4 \, \text{Li}_2\left(x^2\right)$\\[0.3em]
              \hline     
$\begin{array}{l}
4 \left(x^2+1\right)\otimes x\otimes x+24 (1-x)\otimes x\otimes x+12 x\otimes (1-x)\otimes x\\[-1em]+12 x\otimes (x+1)\otimes x+24 (x+1)\otimes x\otimes x \end{array}$& $-\text{Li}_3\left(-x^2\right)-6 \text{Li}_2\left(x^2\right) \ln (x)$\\[0.3em]
\hline
     $\begin{array}{l}
       -6 x\otimes \left(16 x^4-4 x^2+1\right)\otimes x-6 x\otimes \left(4 x^2+1\right)\otimes x\\[-1em]-12 \left(4 x^2+1\right)\otimes x\otimes x-12 \left(16 x^4-4 x^2+1\right)\otimes x\otimes x\\[-1em]+12 x\otimes x\otimes x+5 (2 x+1)\otimes x\otimes x\end{array}$& $\begin{array}{l}\text{Li}_2\left(-64 x^6\right) \ln (x)\\[-1em]-5 \text{Li}_3(-2 x)+2 \ln ^3(x)\end{array}$\\[0.3em]  
       \hline
   \end{tabular}}
   \caption{Examples of data pairs $\{\cS_i, F_i\}$ found in the test set for the symbol integration task. We showcase different symbols with 2 or 3 entries which may be non-trivial to integrate by hand. All examples are solved by the transformer.}
   \label{tab:eg2}
   \end{table}

When crafting our test set we check that it does not contain any symbol also found in the training set. We also make sure that symbols which are linked together by scrambles are not found dispersed across the training and test sets, but rather all contained within the same set. Otherwise during training the transformer would have already seen the expected $F_i$ answer corresponding to a similar (albeit different) symbol. A longer discussion of the training and test sets can be found in Appendix~\ref{app:data}.

\bigskip
\noindent
\textbf{Additional pre-processing for logarithms} \quad 
When keeping overall constant terms we have observed that the transformer would suffer from undesired redundancies that would obstruct the final answer. For instance when performing a beam search, retaining the best $N$ guesses, those would usually only differ from each other by constants inside single logarithms. These constants are not intended to be learned, since they are dropped at the level of the symbol. That is, for single logarithms
\begin{equation}
    \cS[\ln(h(x))] = \cS[c \ln(h(x))]
\end{equation}
Therefore we pre-process our data by removing the constants in front of single logarithms.
In addition we simplify products of logarithms, such as $\ln(x) \ln(x^2) \to 2 \ln^2(x)$ to have a canonical form and avoid an additional redundancy.

\bigskip
\noindent
\textbf{Assessing performance} \quad Following the discussion of Section \ref{sec:transfo-dilogs} we will not be able to get an accurate measure of performance during the training. We rather perform a dedicated evaluation where we parse the predicted mathematical expressions and compare them to the expected output. To get the full picture of the model's performance we will compute the symbol of the transformer's guess $F(x)$  and compare it to the input $\cS_{\text{in}}(x)$  
\begin{equation}
    \cS_{\text{in}}(x)  - \cS[F(x)]  \stackrel{?}{=} 0
\end{equation}
where the comparison is made simpler by fully expanding the symbols when possible. Additionally we find it useful to generate expressions using a hypothesis beam, which increases our chances of recovering the correct solution. A beam search of size $N$ will refer to a transformer model outputting the $N$ most probable ``translations" for a given input. If at least one of the predictions turns out to match the expected input then we consider the example to be solved.

\subsubsection{Running the transformer network}
Using the data generation outlined above we craft two types of datasets, one where we scramble the symbols and one where we do not. When scrambling the symbols we generate about 3.5M pairs of symbols and functions. In this dataset, the same function might appear in pairs with different symbols, due to the scrambling.
For our second dataset we generate symbol and function pairs and instead do not scramble the symbol. So in this data set each function appears paired with exactly one symbol. We let the network run with the same parameters and training time as in Section \ref{sec:transfo-dilogs}. 

\begin{figure}[t]
    \centering
    \includegraphics[width=0.7\textwidth]{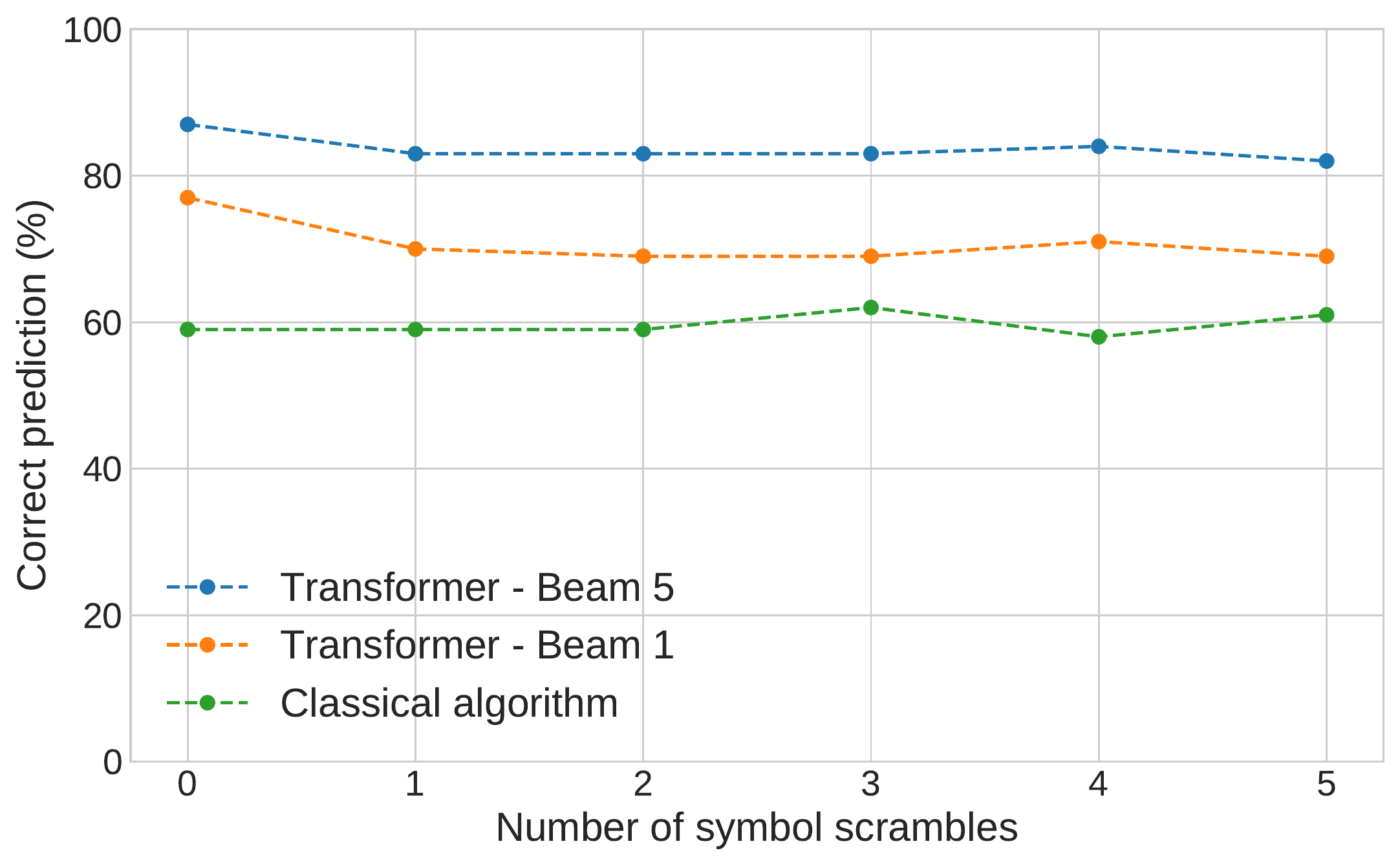}
    \caption{Performance of the transformer network and the classical algorithm on the symbol integration task
    with the dataset of scrambled weight-two symbols.
    For the beam 1 points accuracy is measured by comparing the best guess of the transformer network to the truth. For the beam 5 points, a correct prediction corresponds to any of the network's top 5 guesses being correct.
    }
    \label{fig:scrw2}
\end{figure}

\bigskip
\noindent
\textbf{Solving scrambled symbols} \quad We first look into the 3.5M sample training set where the symbols have been scrambled
and test the performance on an independent sample of 15,000 examples. In this training set, all the functions have transcendental weight 2. 
Performance is evaluated as a function of the number of scrambles and probed with a beam search of size 1 (greedy decoding) or a beam search of size 5 and compared to the classical algorithm detailed in Section \ref{sec:symbol-integration}. Fig.~\ref{fig:scrw2} shows the ability of the transformer network to integrate the symbols efficiently, irrespective of the number of times they have been scrambled. Although there is a minor drop in performance between scrambled and non-scrambled symbols (0 scrambles to $>0$ scrambles in Fig.~\ref{fig:scrw2}), which can be attributed to the length of the input usually going up once we start scrambling the symbol, we can conclude that scrambling the symbol does not seem to impact the model's performance.

\smallskip
\noindent
\textbf{Solving non-scrambled symbols} \quad Since the number of scrambles does not seem to play a large role we can instead focus on integrating the simplified symbol directly. Alternatively we could also consider using a classical algorithm to preprocess the symbol before feeding it to the transformer, since simplifying the symbol is typically much easier than integrating it. We generate three datasets, one with weight two expressions (dilogs), one with weight three expressions (trilogs) and one with weight four expressions, and train separate networks for the three cases. 
Our new training sets have 1.6M, 1.2M and 1M examples respectively. Because of the redundancy in the 3.5M scrambled set (the same function could be paired with multiple symbols) these training sets actually have a larger diversity of functions than those in the 3.5M training set.
 We train separate networks for all experiments (weight 2, weight 3 and weight 4), use test sets of 10,000 examples and list the performance in the Table~\ref{tab:transfoperf}, comparing with a classical algorithm (see Section~\ref{sec:classical_algos}). 

\begin{table}[t]
\centering
\begin{tabular}{l c c c}
\hline
&& \textbf{Beam Size 1}& \textbf{Beam Size 5}\\[0.2em]
 \hline
  \multirow{2}{*}{Weight 2} &Transformer&  82\% &  \textbf{91\%} \\[0.1em]
  &Classical Algorithm & 59\% &  59\% \\[0.1em]
   \hline\\[-1.0em]
Weight 3 & Transformer&  78\% &  \textbf{88\%}  \\[0.1em]
   \hline\\[-1.0em]
Weight 4 & Transformer&  80\% &  \textbf{89\%}  \\[0.1em]
 
\end{tabular}
   \caption{Overall performance on the test set of the symbol integration task where the input symbols have not been scrambled. For the classical algorithm at weight 2 we refer to the section \ref{sec:symbol-integration}. At weight 3 or 4 no classical algorithm is available. A beam size of $N$ refers to testing the $N$ best guesses.}
   \label{tab:transfoperf}
   \end{table}
   
\begin{figure}[ht!]
    \centering
    \includegraphics[width=\textwidth]{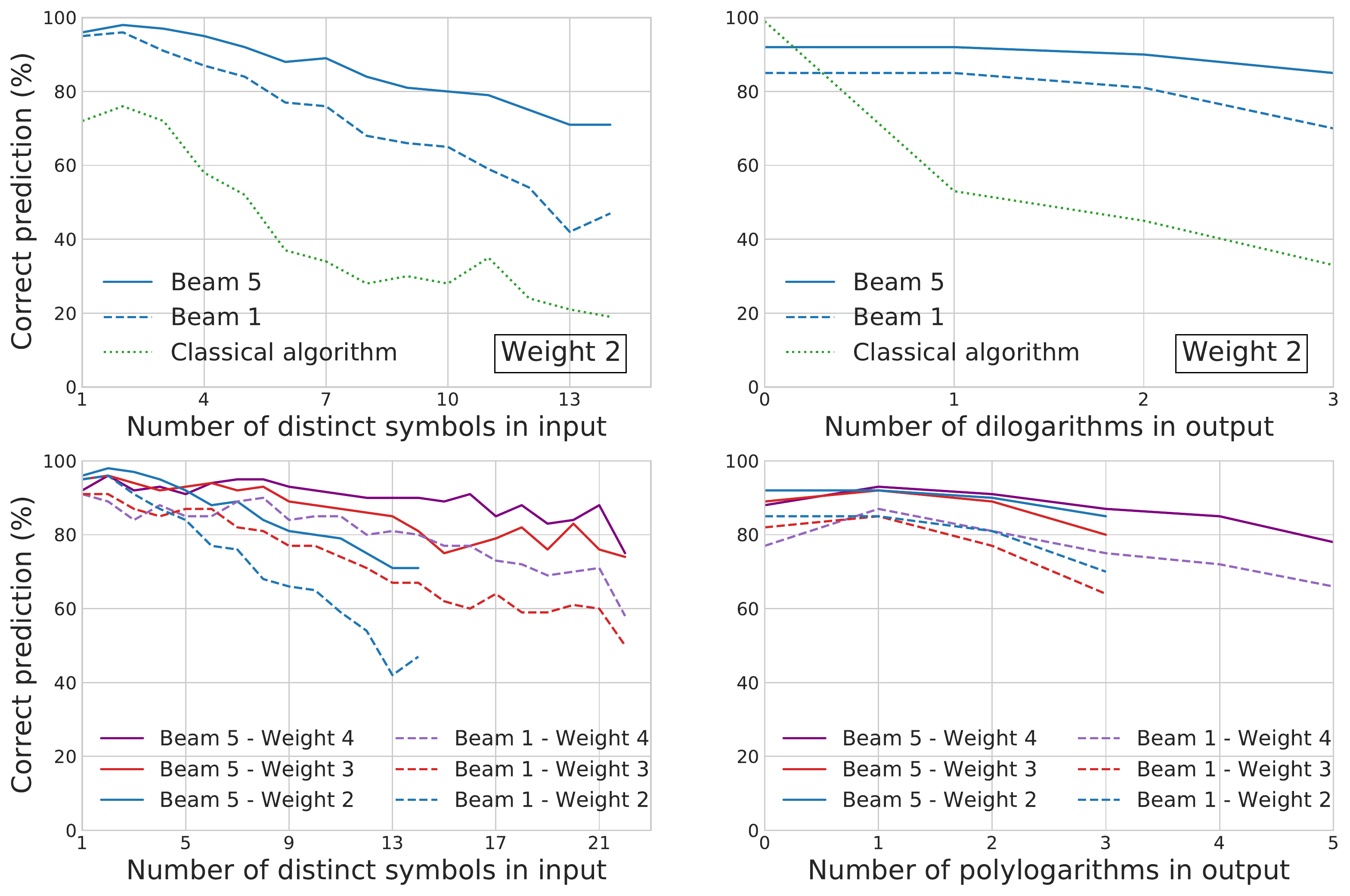}
    \caption{Performance of the transformer in integrating symbols of weight 2 (dilogs), weight 3 (trilogs) or weight 4 ($\Li_4$) as a function of complexity. The input complexity is defined by the number of individual symbol terms in the input expression. The output complexity corresponds to the number of polylogarithms in the simplified expression.  We compare to the classical algorithm only for weight 2, since we did not develop an algorithm for weight 3 or 4.}
    \label{fig:perfint}
\end{figure}

Overall we observe that even with a relatively small dataset (under 2M equations compared to the 40M equations of \cite{fbtransformer}) the transformer is still able to effectively solve the integration task. All functions that appear inside the symbols and in the arguments of the polylogarithms are polynomials, which transformers are able to resolve well \cite{DBLP:journals/corr/abs-1904-01557}. We also note that the performance of the network is largely independent of the transcendentality weight (2, 3 or 4) considered.

One can also get a better insight into the failure modes of the transformer by looking at a measure of the performance as a function of the complexity. We consider the number of symbols in the input prompt and the corresponding number of polylogarithms expected in the simplified form. This analysis is done for our symbol integration task with results displayed in Fig.~\ref{fig:perfint}. We provide a performance figure only if at least 100 relevant examples are found in the test set in order to keep meaningful statistics. For the classical algorithm we consider an integration to be successful irrespective of the length of the solution.
It is apparent from the Fig.~\ref{fig:perfint} that the transformer has a slightly harder time resolving longer sequences, impacting the overall performance. It is important to notice however that long expressions are not found as often in the training data so that the model has less examples to train with. For example at weight 3 close to 80\% of our training examples have less than 14 symbols in the input. The fact that the transformer model can still get close to $80 \%$ accuracy for larger input sequences indicates in fact that global features have been learned and that the network is able to generalize to new data.


\section{Summary example \label{sec:summary}}
Having introduced and trained our transformer network, let us end with an illustrative example of how it might be used, along the lines of the flowchart in Fig.~\ref{fig:flow}. 
Suppose the calculation of a particular scattering amplitude in some quantum field theory resulted in an expression of the form
\begin{multline}
   f(x) =4\zeta_3 +  9 \left[G(0,0,1,x)
   +G\left(0,0,\frac{-1-\sqrt{3}i}{2},x\right)
   +G\left(0,0,\frac{-1+\sqrt{3}i}{2},x\right)\right]\\ 
   +4 \left[-G(-1,-1,-1,x)+G(-1,0,-1,x)+G(0,-1,-1,x)+G(0,0,1,x)-G\left(0,0,1,\frac{x}{x+1}\right)\right]
\end{multline}
Recall that these $G(a,b,c,x)$ are multiple polylogarithms defined in Eq.~\eqref{Gdef}. These functions can be expressed in terms of classical polylogarithms:
\begin{multline}
f(x) = 9 \left(-\text{Li}_3(x)-\text{Li}_3\left(\frac{2 i x}{-i+\sqrt{3}}\right)-\text{Li}_3\left(-\frac{2 i x}{i+\sqrt{3}}\right)\right)\\
+    4 \left(-\text{Li}_3(x)+\text{Li}_3\left(\frac{x}{x+1}\right)+\text{Li}_3(x+1)-\text{Li}_2(-x) \ln (x+1)\right)
\\ - 4 \left(\text{Li}_2(x+1) \ln (x+1)+\frac{1}{6} \ln ^3(x+1)+\frac{1}{2} \ln (-x) \ln ^2(x+1)\right) \,.
\label{fxLi}
\end{multline}
Computing the symbol and simplifying it yields
\begin{equation}
    \cS\big[f(x)\big] = 9 \left(x^2+x+1\right)\otimes x\otimes x+13 (1-x)\otimes x\otimes x+4 (x+1)\otimes x\otimes x
\end{equation}
These steps are all standard. At this point one needs to integrate the symbol, a non-standard and non-trivial task. The transformer network we have trained gives for this symbol the function
\begin{equation}
     -\text{Li}_3(x^3)-\text{Li}_3(x^2)
\end{equation}
Although the network does not predict the transcendental constant, it is straightforward to determine once the functional form is known. The exact result is
\begin{equation}
    f(x) =  -\text{Li}_3(x^3)-\text{Li}_3(x^2)  + 4 \zeta_3
        \label{fxsimp}
\end{equation}
The simplification from Eq.~\eqref{fxLi} to \eqref{fxsimp} is not easy to do classically (try it!) and demonstrates the effectiveness of the machine-learning approach.

\section{Conclusions~\label{sec:conc}}
In this paper we have considered whether machine leaning can help with a mathematical problem essential to many computations of scattering amplitudes in quantum field theory: the simplification of polylogarithmic functions. Polylogarithms $\Li_n(x)$ of weight $n$ are the most basic elements  of a broad class of transcendental functions appearing ubiquitously in physics. Low weight polylogarithms, like $\Li_2(x)$ are known to satisfy a number of identities. However, even for $\Li_2(x)$ knowing which identities to apply in what order to simplify a given expression is a non-trivial task. Much less is known about  identities satisfied by higher weight polylogarithms, making them even more challenging to manipulate and simplify.

We considered first a toy problem where only a fixed set of identities (reflection, inversion and duplication) were allowed on dilogarithmic expressions. In this case, the application of each identity can be thought analogous to moves in a game, making reinforcement learning (RL) a natural machine learning framework. We found that the RL algorithm could effectively learn patterns in the polylogarithms to determine which identity to apply. It performs significantly better (89\% solved) than a random walk (13\% solved) and comparable to a basic classical algorithm (91\% solved). Moreover, the RL algorithm has better scaling than the classical algorithm, requiring fewer exploratory unscrambling steps on average to arrive at the answer. 

The second architecture we considered was that of a transformer network. Transformer networks essentially try to guess the answer, rather than working out a sequence of steps. Thus, while the RL agent will always produce an answer equivalent to the original (but maybe not in the simplest form), the transformer network might guess something that is not actually correct. However, being able to guess the answer allows the transformer network to be more flexible, for example, by not being limited to a finite set of identities input by hand.

We applied the transformer network first to a problem analogous to the RL one: simplifying a dilogarithmic expression. Actually, the transformer's task was somewhat harder than that of the RL in that we asked the transformer network to simplify expressions that did not necessarily simplify to zero. We found similar performance (91\% success rate) on this harder task as the RL network achieved on finding the path to zero.

Then we applied the transformer network in the context of a hybrid algorithm exploiting the power of the symbol. The symbol of a polylogarithmic expression is straightforward to compute, and can be simplified more easily than the original expression. However, converting the symbol back into polylogarithmic form (``integrating the symbol") can be difficult. Thus we trained a transformer network to do this integration step. We found again strong performance: 91\% success for dilogs and 88\% for trilogs. An advantage here of the symbol/transformer approach is that the network seems to work nearly as well with higher weight functions as lower weight ones. In contrast, simplifying expressions involving $\Li_4$ for example is significantly more difficult than simplifying expressions with $\Li_2$ alone.

Although the transformer network we created is not yet powerful enough to be a useful  tool for physics, our results suggest that with additional engineering a powerful tool along these lines is conceivable. 
A practical tool should be able to operate on multivariate functions, for example. In addition, we restricted the arguments to be rational functions with low-degree polynomials and relatively small integer coefficients. Generalizing to arbitrary rational functions would require a much larger state space and becomes difficult to fit into a given network. For example, even though our symbol-integrator transformer network has similar performance on weight 2, 3 or 4 polylogarithmic expressions, the symbols for the higher weight functions are generally much longer and the network training is more challenging and memory intensive. These difficulties can surely be overcome with a more sophisticated architecture, but are beyond our current model.
The networks in this paper are rather a proof-of-concept demonstration of their relevance to symbolic computation in physics.

Concerns about generalization notwithstanding, our networks work better than one might have expected. They can simplify expressions that are challenging to simply by hand, and work (for a narrow task) better than any available symbolic computation package we are aware of. That is not to say that a better classical algorithm  cannot be written -- it most certainly can -- but that a relatively minimal implementation can already perform admirably. One can imagine use cases, such as the generation of new identities among higher-weight polylogarithms, where machine learning could contribute genuinely new results. More generally, it is easy to envision machine learning playing an increasingly larger role in symbolic calculations in theoretical physics. This work can be viewed as a baby step in that direction.


\section*{Acknowledgements}
The computations in this paper were run on the FASRC Cannon cluster supported by the FAS Division of Science Research Computing Group at Harvard University. This research is supported by the National Science Foundation under Cooperative Agreement PHY-2019786 (The NSF AI Institute for Artificial Intelligence and Fundamental Interactions, \url{http://iaifi.org/}).

\appendix
\section{Performance overview of the RL agents} \label{app:perf}
We present here the full performance overview of the RL agents introduced in the Section~\ref{sec:RL}. Represented are the agents with a Graph Neural Network (GNN) in the embedding layer but also those with the simple one-hot encoding scheme. Additionally we make the distinction between the different reward schemes, whereas in the Figure~\ref{fig:allsolved} only the best GNN agents were shown.

\begin{figure}[htp!]
    \centering
    \includegraphics[width=1\textwidth]{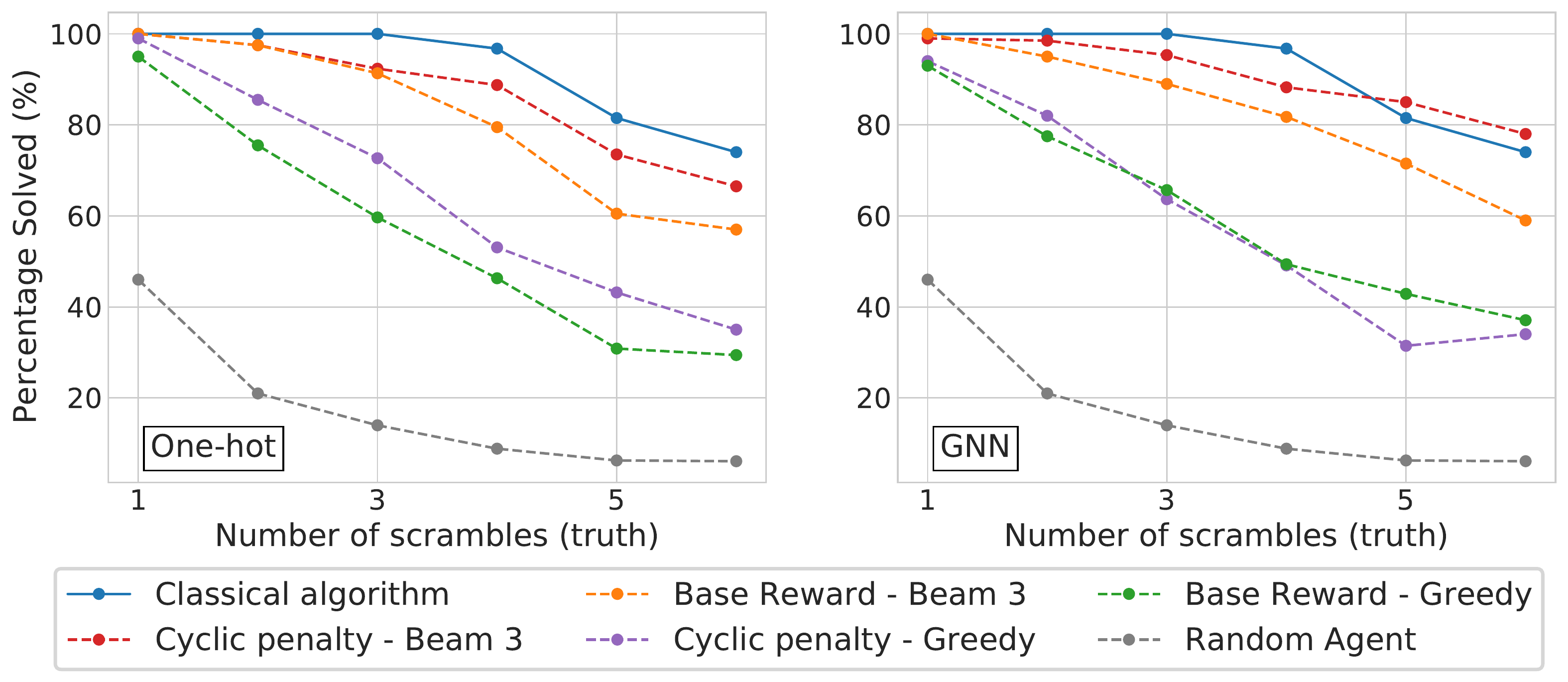}
    \caption{Performance of the different agents on the test set of 1,500 dilogarithmic expressions that simplify to 0. We represent performance as a function of the complexity of the input, given by the number of identities (scrambles) applied to generate the expression.}
    \label{fig:allsolved2}
\end{figure}

\begin{figure}[htp!]
    \centering
    \includegraphics[width=1\textwidth]{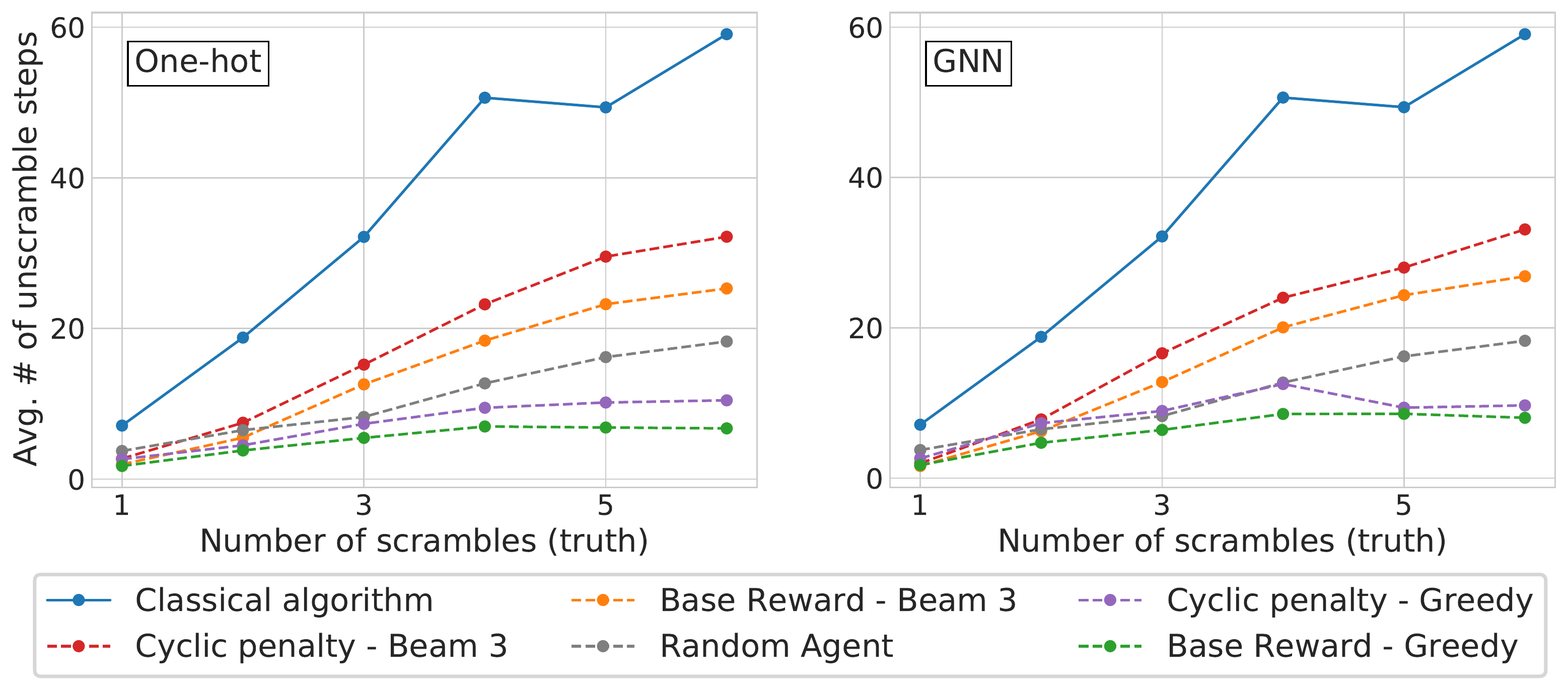}
    \caption{Average number of identities tried by the different agents when solving the examples of the test set. The identities considered here are duplication, inversion or reflection.}
    \label{fig:allnodes2}
\end{figure}

\section{Data generation parameters}\label{secap:data}
For the symbol integration task with transformers we started by generating a set of functions $\{F_i\}$ for which we would then compute the associated symbol set $\{\cS_i\}$. When generating these functions we choose the parameters of Table~\ref{tab:params}.

 \begin{table}[hbtp!]
 \centering
\begin{tabular} {|c|c|c|}
\hline
   Parameter& Range or value & Sampling \\[0.2em]
 \hline
  Distinct terms $n_s$ & [1, 3] & Uniform \\ [0.4em]
 Coefficients $c_j$  & $\left\{a_1, \dfrac{1}{a_2}, \dfrac{a_1}{a_2}\right\}$ & $p=\{0.5, 0.25,0.25\}$  \\[0.8em]
  Coefficients $a_1, a_2$  & [-5,5]&Uniform \\[0.2em]
 \hline
 Unary operator & \{pow2, pow3\}& $p=\{0.2, 0.8\}$  \\[0.2em]
  Binary operator & \{add, sub, mul, div\}& $p=\{0.31, 0.31, 0.15,  0.23\}$  \\[0.2em]
 Number of operators& [0,8] & Uniform \\[0.2em]
 Leaf type& $\{x, const\}$ & $p=\{0.75, 0.25\}$ \\[0.2em]
  Integer leafs & [1,5]& Uniform\\[0.2em]
  \hline
    Weight 2 basis &$\{\Li_2(x), \ln(x)\ln(y)\}$ &$p=\{2/3, 1/3\}$\\[0.4em]
        Weight 3 basis &
     $\left\{  \begin{array}{c}
             \Li_3(x), \Li_2(x)\ln(y),  \\
           \ln(x)\ln(y) \ln(z)
        \end{array}\right\}$
         &$p=\{1/2, 1/3, 1/6\}$\\[1.2em]
                 Weight 4 basis &
     $\left\{  \begin{array}{c}
             \Li_4(x), \Li_3(x)\ln(y),  \\
           \Li_2(x)\Li_2(y)\\
           \Li_2(x)\ln(y)\ln(z)\\
           \ln(w) \ln(x)\ln(y)\ln(z)\\
        \end{array}\right\}$
         &$p=
         \left\{  \begin{array}{c}4/13, 3/13, 3/13 \\ 2/13, 1/13
        \end{array}\right\}$\\[2.5em]
           Repeating argument in basis &$\{\text{Yes},\text{No}\}$ &Uniform\\[0.2em]
  \hline
\end{tabular}
\caption{Set of parameters used for generating the functions in the symbol integration task.}
\label{tab:params}
\end{table}

\noindent
Allowing for the arguments to be repeated in the basis functions grants us access to functions like $\ln(x)^3$ which have been hard to sample otherwise. We also skew the data generation procedure towards producing dilogarithms and trilogarithms since those functions are usually the ones of interest, where integrating the symbol is a challenging task.

\section{Comparing training and testing sets}\label{app:data}

To get a better sense of the learning done by the transformer we further investigate the nature of our training and test sets. In particular one may be worried about potential ``leaks" in the data, as highlighted in \cite{rewriting}, where expressions in the training and test sets differ only by constants factor. To estimate how this could impact our datasets we perform a similarity analysis. We replace all constant factors, including their sign, in the inputs of the training and test sets by a generic token ``$C$" and evaluate how many examples from the test set are truly unique. In the Table~\ref{tab:unique} we report for each of our sets the fraction of examples from the test set that did not map to any example in the training set.

\begin{table}[hbt!]
\centering
\begin{tabular}{c c}
\hline
Dataset & Unique examples in test set (\%)\\[0.2em]
 \hline
  Function simplification of section \ref{sec:transfo-dilogs}& 92\% \\[0.2em]
Scrambled symbol integration (weight 2)& 84\% \\[0.2em]
Non-scrambled symbol integration (weight 2)& 56\% \\[0.2em]
Non-scrambled symbol integration (weight 3)& 66\% \\[0.2em]
Non-scrambled symbol integration (weight 4)& 66\% \\[0.2em]
\end{tabular}
\caption{Similarity analysis between the training and test sets for the various transformer datasets. An example from the test set is unique if it does not match to any example in the training set once all of the constants have been replaced by a generic token ``C".}
\label{tab:unique}
\end{table}

The datasets that were generated with additional diversity have test sets that are clearly separated from the training sets. For the symbol integration task this separation is less present when dealing with non scrambled symbols but might not be as meaningful as for the function simplification task. Indeed, although only differing by constants, a symbol like $(1+3 x^2)\otimes(1+2x^2)$ will integrate to a very different function as compared to a symbol like $(1+2 x^2)\otimes(1+2x^2)$. One might also worry about additional leakage coming from permutations in the inputs, for instance in the function simplification task where we could see $2 \, \Li_2(x) + \Li_2(1-x)$ in the training set and $\Li_2(1-x)+2 \, \Li_2(x)$ in the test set. Our base rational functions are of degree 3 with coefficients in $[-2,2]$. There are about 7000 unique valid functions of the sort, which already implies that we will encounter the same linear combination of two arguments during the generation every $\sim 10^7$ examples (some arguments could be generated in equivalent forms). Our datasets have under 2M examples, so generically this will not happen.

\printbibliography 

\end{document}